%% file: main.tex
\documentclass{article}

\usepackage[preprint]{neurips_2025}

\usepackage[utf8]{inputenc} 
\usepackage[T1]{fontenc}    

\usepackage{xcolor}
\definecolor{ForestGreenCite}{RGB}{34,139,34}
\definecolor{EmeraldGreenLink}{RGB}{80,200,120}
\definecolor{DeepDarkGreenURL}{RGB}{0,60,0}

\usepackage[colorlinks=true, 
            citecolor=DeepDarkGreenURL, 
            linkcolor=EmeraldGreenLink, 
            urlcolor=ForestGreenCite]{hyperref}

\usepackage{url}            
\usepackage{booktabs}       
\usepackage{amsfonts}       
\usepackage{nicefrac}       
\usepackage{microtype}      
\usepackage[a-2u]{pdfx}     

\usepackage[ngerman,american]{babel}
\usepackage[autostyle]{csquotes}

\usepackage{graphicx}
\usepackage{listings}
\usepackage{lstautogobble}
\usepackage{tikz}
\usetikzlibrary{trees, positioning}
\usepackage{pgfplots}
\usepackage{pgfplotstable}
\usepackage{caption}
\usepackage{subcaption}
\usepackage[printonlyused]{acronym}
\usepackage{wrapfig}
\usepackage[justification=raggedright,singlelinecheck=false]{caption}

\usepackage{multirow}
\usepackage[table]{xcolor}
\usepackage{makecell}
\usepackage{longtable}
\usepackage{booktabs}
\usepackage{geometry}
\usepackage{enumerate}

\usepackage{amsmath}
\usepackage{amssymb}
\usepackage{bm}
\usepackage{siunitx}
\usepackage{pifont}

\usepackage{amsthm}
\newtheorem{theorem}{Theorem}[section]
\newtheorem{lemma}[theorem]{Lemma}
\newtheorem{corollary}[theorem]{Corollary}
\theoremstyle{definition}

\theoremstyle{remark}

\newtheorem{claim}[theorem]{Claim}

\usepackage{algorithm2e}

\usepackage{ifthen}
\usepackage{cleveref}
\usepackage{enumitem}
\usepackage{float}
\usepackage{rotating}
\usepackage{todonotes}
\usepackage{marvosym}

\newcommand{\Var}{\operatorname{Var}}
\newcommand{\Cov}{\operatorname{Cov}}
\newcommand{\E}{\mathbb{E}}

\title{Tree-OPO: Off-policy Monte Carlo Tree-Guided Advantage Optimization for Multistep Reasoning}

\author{
  Bingning Huang\textsuperscript{1,}\thanks{Equal contribution.} \thanks{Interned at the AI4Sec team, Huawei Munich Research Center.} \quad
  Tu Nguyen\textsuperscript{2,}\footnotemark[1] \quad
  Matthieu Zimmer\textsuperscript{3} \\
  \\
  \textsuperscript{1}Technical University of Munich \\
  \textsuperscript{2}Huawei R\&D Munich \\
  \textsuperscript{3}Huawei Noah’s Ark Lab \\
  \texttt{bingning.huang@tum.de} \quad
  \textsuperscript{\Letter}\{\texttt{tu.nguyen}, \texttt{matthieu.zimmer}\}\texttt{@huawei.com}
}

\begin{document}

\maketitle

\begin{abstract}
Recent advances in reasoning with large language models (LLMs) have shown the effectiveness of Monte Carlo Tree Search (MCTS) for generating high quality intermediate trajectories, particularly in math and symbolic domains. Inspired by this, we explore how MCTS derived trajectories, traditionally used for training value or reward models, can be repurposed to improve policy optimization in verifier guided reinforcement learning (RL). Specifically, we focus on Group Relative Policy Optimization (GRPO), a recent algorithm that enables consistent policy learning from group relative judgments. We reframe GRPO into a staged training paradigm, leveraging a teacher's MCTS rollouts to construct a tree structured curriculum of prefixes. This introduces the novel challenge of computing advantages for training samples that originate from different prefixes, each with a distinct expected return. To address this, we propose Staged Advantage Estimation (SAE), a framework for computing low variance, prefix aware advantages by projecting rewards onto a constraint set that respects the tree's hierarchy. Our empirical results on mathematical reasoning tasks show that SAE improves final accuracy over standard GRPO. This outcome is grounded in our theoretical analysis, which confirms that SAE reduces gradient variance, a principled path to improved sample efficiency. We demonstrate this through practical SAE implementations, comparing efficient heuristics against a formal quadratic program.
\end{abstract}

\section{Introduction}
\label{sec:intro}
Many reasoning-intensive tasks, from mathematical problem-solving to symbolic analysis in security domains like deobfuscation and reverse engineering, require multi-hop compositional inference~\citep{christianos2023pangu}. While Monte Carlo Tree Search (MCTS) has been widely used in structured decision-making (most notably in game-playing~\citep{alphago}), recent work has shown its utility in language-based reasoning as well~\citep{rstar, omegaPRM, mathshepherd, rest-mcts}. In these settings, MCTS systematically explores the reasoning space and balances exploration with exploitation, enabling it not only to improve test-time inference but also to identify reliable reasoning chains. These high-quality trajectories are commonly used to supervise downstream training, typically by training reward models~\citep{rstar, omegaPRM, mathshepherd} or enabling self-training from model-generated traces~\citep{10.1145/3726302.3729965, rest-mcts}. In some cases, a strong teacher model guides MCTS rollouts, effectively distilling reasoning behavior into downstream models via these enriched trajectories~\citep{jiang2024enhancing}.

We draw inspiration from this line of work, but take a distinct path: rather than using MCTS \textit{rollouts} solely for supervised reward modeling or test-time inference, we repurpose \textit{partial trajectories}, generated by a strong teacher policy, to construct enriched training groups that drive policy learning. This places emphasis on learning from the comparative quality of multiple sampled behaviors instead of estimating a scalar value. In this context, we examine Group Relative Policy Optimization (GRPO) \citep{deepseekr1}, which performs policy updates using group-level ranking signals rather than relying on an explicit value function.

Our main observation is that the staged nature of MCTS rollouts naturally induces a \textbf{tree-structured space of prompt-completion pairs}~\footnote{We use "staged" and "tree-structured" interchangeably for this hierarchical setting.}, where each path from root to leaf corresponds to a sequence of reasoning steps. This setting introduces a previously unstudied challenge in GRPO: how to compute meaningful advantages when group samples correspond to different prefixes in a shared trajectory. To address this, we introduce \emph{Staged Advantage Estimation (SAE)}, a method for computing advantages across this structured curriculum by accounting for prefix-conditioned values.

We evaluate the staged GRPO setting on mathematical reasoning tasks, currently training primarily on GSM8K~\citep{gsm8k}, and observe measurable improvements over standard GRPO. These gains arise from our SAE, which explicitly exploits the prefix-conditioned structure of MCTS rollouts to compute more informative advantage estimates. Theoretical analysis shows that stage-wise baselines reduce gradient variance and align with conditional success probabilities, providing principled guidance for advantage computation. To make SAE practical, we introduce heuristic approximations that balance bias and variance: they stabilize updates while avoiding the need for extensive rollouts per prefix, which would otherwise be required for low-variance subtree estimates. While SAE performs robustly, the magnitude of improvements is naturally limited by objective factors such as model capacity and dataset difficulty, suggesting that richer tree-structured supervision could yield larger benefits on more challenging mathematical reasoning tasks (than GSM8K). Our work highlights new directions in integrating structured trajectory supervision with lightweight RL methods and opens up theoretical questions about policy learning in tree-structured domains.

\section{Related Works}
LLM reasoning and distillation has been widely explored in literature, though often in isolation. Approaches like MiniLLM \citep{minillm}, DistiLLM~\citep{distillm, distillm2} and GKD \citep{gkd} focus on general distillation, typically leveraging Kullback-Leibler (KL) divergence on logits to transfer universal capabilities or factual knowledge. However, these methods often lack a direct focus on distilling intricate reasoning processes, instead concentrating on broader knowledge transfer. More specialized distillation efforts, such as ~\citep{taskd-llm, distill-step-by-step}, aim for task-specific knowledge transfer, with the latter even targeting intermediate thought processes (rationales). While these works delve into transferring knowledge about \textit{how} to reason (e.g., specific steps), they primarily focus on reproducing teacher-generated content rather than transferring the underlying \textit{skill} of exploring and evaluating reasoning paths.

Separately, reasoning-based RL paradigms, such as GRPO~\citep{grpo} and particularly, its tree-based variants~\citep{treerpo, treepo}, address complex problems by optimizing policies over tree-structured action spaces. However, these methods typically operate in an online, on-policy setting, requiring full-trajectory interactions to compute advantages. In contrast, our approach operates in a unique hybrid setting. We leverage partial trees generated offline by a teacher's policy (e.g., via MCTS) and then use the student policy to complete these sub-trajectories online. This curriculum-driven process results in a \textit{novel tree-structured advantage space} where reasoning is studied without requiring full, costly trajectory interactions. To our knowledge, our work represents the first comprehensive exploration into improving complex reasoning skills in LLMs by actively studying the structure of the advantage space beyond flat or simple group-wise considerations. This offers a novel approach to cultivate sophisticated reasoning abilities.

\section{Methodology}

We extend GRPO, a method that relies on \textit{on-policy} rollouts where the target policy must rediscover entire reasoning paths from a single prompt. This standard approach can be sample-inefficient for complex tasks due to sparse rewards and challenging exploration. Our key insight is to replace this on-policy sampling with a more structured, \textit{off-policy} curriculum. We use a strong teacher policy to perform MCTS offline, generating a rich dataset of valuable partial trajectories (prefixes). The student then learns by starting its rollouts from these informative, teacher-vetted reasoning states.

This reframes the learning problem from solving a single hard task to mastering a curriculum of diverse, more tractable sub-problems. By decomposing each MCTS rollout into a tree of prefixes, we achieve significant variance reduction and provide a denser, more targeted learning signal, yielding our \textit{Tree-structured Off-policy Optimization} (\textbf{Tree-OPO}) framework.

\subsection{Problem Setting and  MCTS‑Derived Prefix Tree}

Consider a reasoning task modeled as a Markov decision process (MDP) \((\mathcal{S}, \mathcal{A}, T, r)\), where each state \(s \in \mathcal{S}\) is a prefix \(p = (t_1, \dots, t_n)\)—a sequence of tokens representing a partial solution—and each action \(a \in \mathcal{A}\) generates the next token. Transitions are deterministic: \(T(p' = (p \Vert a) \mid p, a) = 1\) where $p \Vert a$ is the concatenation of the prefix with the new token, and rewards \(r(p) \in \{0, 1\}\) are provided only at terminal completions based on task success. Our objective is to learn a policy \(\pi_\theta(a \mid p)\) that, given an initial prompt \(p^{(0)}\), auto-regressively generates a correct full solution.

\paragraph{Offline teacher‐generated prefixes.}  

\begin{wrapfigure}{r}{0.44\columnwidth}
  \vspace{-1em}
  \centering
\resizebox{0.4\columnwidth}{!}{%
    \begin{tikzpicture}[
      level distance=1.5cm,
      sibling distance=2.5cm,
      node/.style = {circle, draw, minimum size=1cm, align=center, font=\small},
      edge from parent/.style = {draw, -latex},
      hard/.style   = {fill=red!30},
      medium/.style = {fill=yellow!40},
      easy/.style   = {fill=green!30}
    ]

    \node[node,hard] {Q \\ (0)}
      child {node[node,medium] {A \\ (1)}
        child {node[node,medium] {B \\ (2)}
          child {node[node,easy] {C \\ (3)}
            child {node[node,easy, label={[label distance=-4pt]below:{\rotatebox{-90}{~~$\sim\sim\!>$}}}] {D \\ (4)}} 
          }
        }
        child {node[node,easy] {J \\ (2)}
          child {node[node,easy, label={[label distance=-4pt]below:{\rotatebox{-90}{~~$\sim\sim\!>$}}}] {K \\ (3)}} 
        }
      }
      child {node[node,medium] {E \\ (1)}
        child {node[node,easy] {F \\ (2)}
          child {node[node,easy, label={[label distance=-4pt]below:{\rotatebox{-90}{~~$\sim\sim\!>$}}}] {G \\ (3)}} 
        }
      }
      child {node[node,medium] {H \\ (1)}
        child {node[node,easy, label={[label distance=-4pt]below:{\rotatebox{-90}{~~$\sim\sim\!>$}}}] {I \\ (2)}} 
        child {node[node,easy, label={[label distance=-4pt]below:{\rotatebox{-90}{~~$\sim\sim\!>$}}}] {L \\ (2)}} 
      };

    \end{tikzpicture}
  }
  \caption{\textit{Staged‐reasoning tree with \emph{reverse} curriculum coloring: deeper prefixes are easier (green); shallow prefixes are harder (yellow). Arrows at leaves denote completing trajectories.}}
  \label{fig:MCTS_tree_curriculum}
\end{wrapfigure}
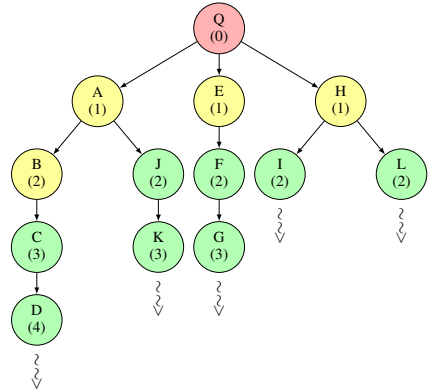

An expert teacher—equipped with a stronger model—runs MCTS offline to produce full solution traces
\[
  \tau = \bigl(p^{(0)}, a^{(0)}, a^{(1)}, \dots, a^{(k)}\bigr).
\]
We decompose each trace into staged prefix–completion pairs where each \(p^{(i)}\) is the prefix after \(i\) steps:
\[
  \bigl\{(p^{(i)},\,c^{(i)})\bigr\}_{i=0}^{k}, 
  \quad
  c^{(i)} = \bigl(a^{(i)}, a^{(i+1)}, \dots, a^{(k)}\bigr),
\]
and collect all prefixes \(p^{(i)}\) into a directed‐acyclic reasoning tree (Fig.~\ref{fig:MCTS_tree_curriculum}).  While the student policy could, in principle, run MCTS online~\citep{treepo, treerpo}, we leverage this precomputed teacher dataset to minimize compute and capture expert trajectories without additional extensive search. Details on the staged rollout and teacher-generated prefixes are in Appendix~\ref{app:teacher_prefixes}.

\paragraph{Online Tree-OPO rollouts.}
At each update, we sample a minibatch of $K$ prefixes $\{p_k\}$ from the teacher-generated tree, according to a fixed distribution $\rho_D$. The student policy $\pi_\theta$ then generates continuations $\hat c_k\sim\pi_\theta(\cdot\mid p_k)$, forming a complete online trace $\tau_k = p_k \Vert \hat c_k$. For each $\tau_k$, we receive a binary reward $r_k \in \{0,1\}$ for full-problem success. The learning objective is to maximize the expected reward on the distribution of prefixes provided by the teacher:
\[
L(\theta) = \mathbb{E}_{p \sim \rho_D, \hat c \sim \pi_\theta(\cdot \mid p)}[r(p \Vert \hat c)].
\]

\noindent We form groups of \((\tau_k,r_k)\) and compute the policy gradient using a refined advantage signal, \(A_k\), which is obtained through our staged advantage estimation (SAE) method (Section~\ref{sec:SAE})

\noindent
Optionally, when a teacher policy $\pi_T$ is available, we can correct for off-policy sampling bias by introducing an importance weighting term. Following the generalized soft policy optimization (GSPO) framework~\citep{gspo}, the objective can be written as
\begin{equation}
\label{eq:gspo_importance}
\mathcal{L}(\theta)
=
\mathbb{E}_{p \sim \rho_D,\, \hat c \sim \pi_T(\cdot \mid p)}
\!\left[
w_\theta(p,\hat c)\, r(p,\hat c)
\right],
\qquad
w_\theta(p,\hat c)
=
\frac{\pi_\theta(\hat c \mid p)}{\pi_T(\hat c \mid p)}.
\end{equation}
This formulation reweights teacher-guided rollouts $\hat c$ under the teacher-induced prefix distribution, ensuring that $\nabla_\theta \mathcal{L}(\theta)$ remains an unbiased estimator of the on-policy objective of $\pi_\theta$.
When combined with staged advantage estimation (SAE), the importance ratio provides a principled mechanism to integrate the teacher’s sampling distribution with the student’s optimization dynamics, yielding either sequence- or token-level variants analogous to GSPO-token~\citep{gspo}, while maintaining numerical stability under off-policy sampling.

\paragraph{Reverse curriculum and Tree-OPO}  
In the MCTS-generated prefix tree, deeper prefixes—those with more context—tend to yield higher success rates, as they represent easier subproblems. Although our prefix sampling is uniformly random, this structural property induces an implicit \emph{reverse curriculum}: sampled prefixes naturally span a range of difficulties, from simple continuations (deep nodes) to full-problem completions (shallow nodes). Tree‑OPO exploits this variation by organizing on-policy rollouts into depth-mixed groups, allowing the policy to learn from both low- and high-difficulty subproblems within each update. This tree-structured decomposition promotes stability and improves sample efficiency compared to flat trajectory methods.

\subsection{Staged Advantage Estimation as Constrained Quadratic Program}
\label{sec:SAE}

In standard GRPO, a single baseline (the group mean) suffices because all completions share the same prompt context. However, in our Tree‑OPO setting each sample in a group may originate from a different prefix \(p_k\) of the MCTS-generated reasoning tree, and these prefixes have {\em disparate} expected returns. Without correcting for this, shallow prefixes (low \(\E[r\mid p]\)) and deep prefixes (high \(\E[r\mid p]\)) are unfairly compared, leading to:

\begin{itemize}[nosep]
  \item \textbf{Misaligned baselines:} Using a single mean ignores each prefix’s own expected return \(\E[r\mid q_i]\), leading to biased advantages.
  \item \textbf{Increased variance:} Mixing prefixes of different depths without context-specific centering amplifies gradient noise and hinders convergence.
  \item \textbf{Erroneous credit assignment:} Without tree‑aware baselines, a strong outcome on an easy (deep) prefix can overshadow genuine improvements on a harder (shallow) prefix, and v.v.
\end{itemize}

To address these issues, we formulate advantage estimation as the solution of a constrained quadratic program over the group of staged samples. For a group of \(N\) staged pairs \(\{(p_k,r_k)\}\) with binary rewards \(r_k\in\{0,1\}\), let $\bm{a} = (a_1, \ldots, a_N)^T$ be the vector of advantages and $\bm{r} = (r_1, \ldots, r_N)^T$ be the vector of rewards. The optimization problem is:

\begin{equation}
\begin{aligned}
\min_{\mathbf a\in\mathbb R^N}\quad & \;\|\mathbf a - \mathbf r\|^2,\\[-2pt]
\text{s.t.}\quad
& \; \mathbf 1^\top \mathbf a = 0,\\[-2pt]
& \;\|\mathbf a\|^2 \le N,\\[-2pt]
& \; a_i + \delta_{ij} \le a_j \quad\forall\,(i,j)\in\mathcal C_{\mathrm{order}},
\end{aligned}
\label{eq:sae-opt}
\end{equation}

The constraints in Eq.~\ref{eq:sae-opt} can be configured in two primary modes. For practical application, we use a \textit{hard}-constraint mode, which may enforce a strict non-convex normalization $\|\bm a\|^2 = N$ and positive margins $\delta_{ij} > 0$ to produce a more discriminative advantage signal. For our theoretical analysis, we define a \textit{soft}-constraint mode, which uses a convex relaxation of these constraints ($\|\bm a\|^2 \le N$ and $\delta_{ij} = 0$) to guarantee a unique, well-behaved solution. While our theory confirms the desirable properties of this soft formulation, our current experiments focus on the hard-constraint variant; a direct empirical comparison is deferred to future work. The final set of ordering constraints, $\mathcal{C}_{\mathrm{order}}$, is common to both modes and encodes the crucial tree-consistency inductive bias based on relationships within the MCTS tree.

To formally define these constraint sets, let $\mathcal{D}$ denote the entire set of MCTS-generated traces for the current group. We introduce the following predicate functions:
\begin{itemize}[nosep]
    \item $\textit{IsPrefix}(i,j)$: True if $p_i \subset p_j$.
    \item $\textit{IsSibling}(i,j)$: True if $p_i$ and $p_j$ share the same immediate parent prefix. 
   \item $\textit{HasSuccessfulContinuation}(p_k)$: True if an online rollout $\hat{c}$ from $\pi_\theta$ forms a complete, successful trace $\tau_m = p_k \cdot \hat{c}$ (i.e., its binary reward $r_m=1$). For brevity, we denote this as $S(p_k)$ or $S(k)$ for trace $p_k$.
\end{itemize}

\paragraph{Pair-wise (Parent-Child) Consistency Constraints ($\mathcal{C}_{\text{pair}}$)}
These constraints ensure advantages respect basic prefix containment when a failing prefix is extended into a successful trace:
$$ \mathcal{C}_{\text{pair}} = \{ (i,j) \mid \textit{IsPrefix}(i,j),\; r_i=0,\; r_j=1 \}. $$

\paragraph{Triplet Consistency Constraints ($\mathcal{C}_{\text{triplet}}$)}
These constraints refine advantage ordering among sibling prefixes based on more nuanced observations of their future potential via the current policy's online rollouts. Specifically, for sibling prefixes $p_i$ and $p_j$ that both yield immediate failed online rollouts ($r_i=0, r_j=0$) and have not yet exhibited successful completions from the student's online policy ($\neg S(p_i),\; \neg S(p_j)$), we enforce $a_j > a_i$ if $p_i$ is a prefix of a deeper path $p_k$ for which a successful completion \textit{has} been observed ($S(p_k)$). This prioritizes exploration of branches that are currently \textit{less proven} in their overall success trajectory.

$$\mathcal{C}_{\text{triplet}} = \{ (i,j,k) \mid \textit{IsSibling}(i,j),\; r_i=0,\; r_j=0,\; \neg S(p_i),\; \neg S(p_j),\; \textit{IsPrefix}(i,k),\; S(p_k) \}.$$

\begin{figure}[t]
  \centering
  \includegraphics[width=0.9\linewidth]{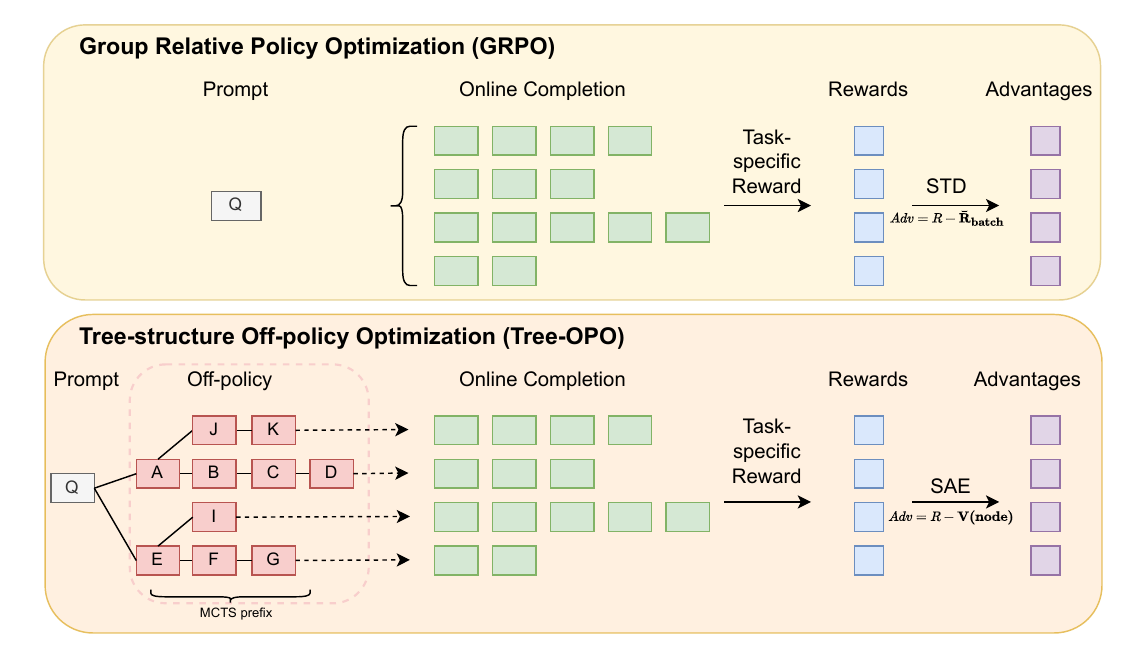}
  \caption{\textbf{Tree-OPO vs.\ GRPO.} \textit{GRPO (top) standardizes rewards from single-prompt completions, inherently limits its ability to differentiate advantages across paths stemming from common prefixes with disparate expected returns. In contrast, our Tree-OPO (bottom) is designed to adhere to Equation (\ref{eq:sae-opt}), facilitating hierarchical advantage ordering aligned with the tree-induced curriculum. This replaces flat standardization, yielding more discriminative advantages for structured generation.}}
  \label{fig:sae}
\end{figure}

\subsection{Heuristic Baselines for Advantage Calculation}

While SAE is formally defined by the constrained quadratic program in Eq.~\eqref{eq:sae-opt}, solving this for every training batch can be computationally intensive. We therefore primarily evaluate several computationally efficient \textbf{heuristic baselines} (e.g., expectation, optimistic) that approximate the prefix-conditioned advantages.\footnote{Normalizing by the standard deviation, as is common in policy-gradient methods, would rescale advantages inconsistently across prefixes and introduce bias into stage comparisons; see App.~\ref{app:baseline-normalization}.} To benchmark these heuristics against the formal SAE definition, we also evaluate the \textbf{SAE (Hard Constraints)} method, which uses the non-convex \textit{hard} formulation ($\|\bm a\|^2 = N$). An empirical comparison against the \textit{soft} formulation~\ref{eq:sae-opt} used in our theoretical analysis is deferred to the extended version.

For each sampled prefix $p_i$ with reward $r_i$, we first compute a raw advantage:
\[
a'_i = r_i - \alpha\, V(p_i), \quad \alpha \in [0,1],
\]
where $V(p)$ is a heuristic estimate of the expected return from prefix $p$, summarizing outcomes of rollouts within its subtree. These raw advantages $a'_i$ are then mean-centered to obtain the final advantages $a_i$, ensuring zero-mean and stabilizing training by reducing bias in gradient estimates.

We consider three heuristic baseline estimators for $V(p)$:

\begin{enumerate}[nosep]
  \item \textbf{Empirical (Expectation):}  
    \[
      V_E(p) = \tfrac{\#\{\text{successful rollouts from } p\}}{\#\{\text{total rollouts from } p\}},
    \]
    the subtree success rate. By Lemma~\ref{lem:exp}, the ideal baseline
    \(V^*(p)=\mathbb{E}[r\mid p]\) uniquely minimizes variance and maximizes reward-alignment. $V_E(p)$ is its natural Monte--Carlo (MC) approximation. Combined with mean-centering,
    \(
      a'_i = r_i - \alpha V_E(p_i),
    \)
    this yields tree-consistent advantages with near-optimal variance reduction.

  \item \textbf{Optimistic:}  
    \[
      V_O(p) = \mathbf{1}\{\exists\,\text{successful rollout in $p$'s subtree}\},
    \]
    which amplifies sparse positive signals and encourages exploration.

  \item \textbf{Pessimistic:}  
    \[
      V_P(p) = \mathbf{1}\{\nexists\,\text{failed rollout in $p$'s subtree}\},
    \]
    promoting conservative updates by penalizing uncertainty in downstream paths.
\end{enumerate}

These heuristics capture different assumptions about future trajectories while remaining tree-consistent and computationally light. Together with mean-centering, they approximate the variance- and reward-alignment properties of the ideal solution to \eqref{eq:sae-opt}, offering a practical substitute for exact constrained optimization. Overall, the Empirical (Expectation) baseline tends to deliver the best balance between variance reduction and reward alignment, while the Optimistic and Pessimistic baselines remain valuable in settings such as sparse rewards or high-stakes pruning, where exploration or conservativeness may be more desirable than strict variance minimization.

\subsection{Monte Carlo Policy-Gradient Interpretation}
Each update samples \(K\) prefixes \(p_k\) from the MCTS-derived tree. The student policy \(\pi_\theta\) then generates full continuations \(\hat c_k\sim\pi_\theta(\cdot\mid p_k)\) and each completed trajectory yields a binary reward \(r_k\in\{0,1\}\). The policy gradient estimator using staged advantages \(A_k = r_k - \alpha V(p_k)\) (Sec.~\ref{sec:SAE}) is
\[
\widehat{g}=\frac{1}{K}\sum_{k=1}^K A_k\,\nabla_\theta\log\pi_\theta(\hat c_k\mid p_k),
\]
where, for autoregressive models, \(\nabla_\theta\log\pi_\theta(\hat c_k\mid p_k)=\sum_{t=0}^{|\hat c_k|-1}\nabla_\theta\log\pi_\theta(c_{k,t+1}\mid p_k,c_{k,0:\,t})\).

The ideal stage baseline of Lemma~\ref{lem:exp} admits the closed form
\[
V(p)=\mathbb{E}[r\mid p]=\Pr_{\pi_\theta}(\text{success}\mid p),
\]
i.e., the success probability of continuations from prefix \(p\). We estimate this with MC rollouts:
\[
\widehat V_M(p)=\frac{1}{M}\sum_{m=1}^M r^{(m)},\qquad r^{(m)}\in\{0,1\},
\] (See Appendix~\ref{app:mc_baseline} for a short derivation, variance formula and simple concentration bounds.)
which is unbiased and satisfies \(\Var[\widehat V_M(p)]=\Var[r\mid p]/M\). Under sparse rewards this variance can be large, creating a computational tradeoff: reducing estimator variance requires many rollouts.

This tradeoff motivates our practical choices in Sec.~\ref{sec:SAE}: mean-centering combined with heuristic baselines (empirical / optimistic / pessimistic) provides a low-cost approximation to \(\mathbb{E}[r\mid p]\) while implicitly encoding tree-consistency. By Theorem~\ref{theo:tree-signal}, the SAE projection of these staged advantages onto the convex set defined by tree-structured order constraints reduces or preserves variance, which directly controls the noise in the MC policy gradient estimator and improves sample efficiency.

\subsection{Theoretical Analysis}
\label{sec:theory}

\paragraph{Key assumptions.}
We make three short, explicit assumptions used in our statements below:
\begin{enumerate}[label=(A\arabic*), nosep]
  \item\label{as:consistency} \textbf{Constraint consistency.} The ordering set $\mathcal C_{\mathrm{order}}$ constructed from MCTS traces is non-contradictory (acyclic). If violated, we use a penalized/soft relaxation.
  \item\label{as:convex}\textbf{Convexified/soft normalization.} For theoretical guarantees we analyze a convexified SAE program (RMS equality relaxed to a convex ball) or its penalized counterpart; this leaves the intuitive constraints intact while producing a closed, (locally) well-behaved feasible set.
  \item\label{as:regularity}\textbf{Policy Gradient Regularity.} The log-policy gradients are uniformly bounded in expectation, a standard condition required to formally analyze the variance of the gradient estimator.
\end{enumerate}

The formal, assumption-aware statements and complete proofs are deferred to Appendix~\ref{appendix:theory}. Below we state the results used in the text and give short proof sketches to clarify the mechanism. The learning objective is the standard expected policy return, $J(\theta) = \mathbb{E}_{\tau \sim \pi_\theta}[r(\tau)]$.

\begin{lemma}[Unbiasedness~\citep{williams1992simple}]
\label{lem:unbias}
Let \( \widehat{g} = \frac{1}{K} \sum_{k=1}^K a_k\,\nabla_\theta \log \pi_\theta(a_k \mid p_k) \), where \( a_k = r_k - \alpha V(p_k) \) and \(V(p)\) is any deterministic function of the prefix. Then
\[
\mathbb{E}_{p,a \sim \pi_\theta}[\widehat{g}] = \nabla J(\theta).
\]
\end{lemma}
\paragraph{Comment.} This standard result confirms that the policy gradient estimate is invariant in expectation to the subtraction of an arbitrary state-dependent baseline $V(p)$. The practical consequence is that the problem of estimator variance can be addressed independently of gradient unbiasedness. This licenses the search for an optimal variance-reducing baseline, which the subsequent lemma provides.

\begin{lemma}[Optimality of Expectation Baseline~\citep{greensmith2004variance}]
\label{lem:exp}
The baseline \(V(p) = \mathbb{E}[r \mid p]\) minimizes variance among deterministic functions of \(p\) and maximizes \(\Cov(r, V(p))\); hence it yields the most reward-aligned advantage signal (formal statement and proof in Appendix~\ref{app-lem:exp}).
\end{lemma}

\paragraph{Comment.} 
The closed-form identity is
\[
V(p)=\mathbb{E}[r\mid p]=\Pr_{\pi_\theta}(\text{success}\mid p),
\]
so one may estimate it by MC rollouts $\widehat V_M(p)=\frac{1}{M}\sum_{m=1}^M r^{(m)}$. In sparse-reward domains the variance of this estimator can be large (variance scales as $1/M$), motivating the heuristic baselines (empirical / optimistic / pessimistic) we use in practice; these heuristics trade bias for lower estimator variance while approximately tracking $\mathbb{E}[r\mid p]$ (Sec.~\ref{sec:SAE}).

\medskip
We next make precise how structural (tree) constraints on advantages encode an inductive bias useful for preference updates.

\begin{lemma}[Tree-Induced Advantage Structure and Inductive Bias]
\label{lem:tree-structure}
Solving the SAE objective (Eq.~\ref{eq:sae-opt}) or its convex/penalized relaxation with ordering constraints $\mathcal{C}_{\mathrm{order}}=\mathcal{C}_{\mathrm{pair}}\cup\mathcal{C}_{\mathrm{triplet}}$ yields an advantage vector $\bm a^*$ that enforces \textbf{prefix-consistent ranking}: for every $(i,j)\in\mathcal C_{\mathrm{order}}$ we have \(a_i^* + \delta_{ij} \le a_j^*\) (with margin~$\delta_{ij}\ge0$ in the convex program).
The precise convexified program and proof are in Appendix~\ref{app-lem:tree}.
\end{lemma}

\paragraph{Comment.} This makes explicit the inductive bias built into SAE: raw rewards are adjusted minimally while obeying the ordering constraints extracted from MCTS traces. The consequence is a set of advantages that respect prefix-consistency, meaning that if one continuation is preferred to another in the tree, the adjusted signal reflects this preference with a margin. The result is not just variance control, but an embedding of structured relational information into the gradient signal itself: triplet constraints re-weight siblings systematically, biasing exploration toward less-proven but promising branches. This inductive bias is the foundation for the guarantees that follow.

\medskip

\begin{theorem}[Tree Constraints Improve Gradient Signal]
\label{theo:tree-signal}
Under the standing assumptions and for a feasible, consistent constraint set, the convex/penalized SAE projection satisfies
\[
\Var[\bm a^*]\le \Var[\bm r],
\]
and enforces positive pairwise margins for constrained pairs, thereby improving group-level discriminability used by preference updates. (Precise technical conditions and proof: Appendix~\ref{app-theo:tree-signal}.)
\end{theorem}

\paragraph{Comment.} This formalizes the intuition that order-enforcing projections act as a variance-reducing filter on the noisy raw rewards. By projecting onto a closed convex set consistent with ordering, the resulting advantages are guaranteed to have no greater variance than the centered rewards. At the same time, the constraints enforce discriminative margins that improve preference alignment at the group level. This establishes that the structured projection not only stabilizes the signal but also sharpens its usefulness for preference-based updates.

\medskip
Beyond variance control, SAE also improves the \emph{estimation-to-class} error of prefix values by enforcing tree-consistent structure.
\medskip

\begin{theorem}[SAE reduces estimation-to-class error]\label{theo:estimation-error-main}
Let $\widehat{\mathbf V}_{\mathrm{GRPO}}\in\mathbb R^P$ be the per-prefix empirical means and
$\mathcal C\subset\mathbb R^P$ the closed, convex set of tree-consistent prefix-value vectors.
Define $\widehat{\mathbf V}_{\mathrm{SAE}}=P_{\mathcal C}(\widehat{\mathbf V}_{\mathrm{GRPO}})$ and
$\mathbf V^*_{\mathcal C}=P_{\mathcal C}(V^\pi)$.
Then
\[
\big\|\widehat{\mathbf V}_{\mathrm{SAE}}-\mathbf V^*_{\mathcal C}\big\|_2
\;\le\;
\big\|\widehat{\mathbf V}_{\mathrm{GRPO}}-\mathbf V^*_{\mathcal C}\big\|_2,
\]
with strict inequality whenever $\widehat{\mathbf V}_{\mathrm{GRPO}}\notin\mathcal C$ and
$P_{\mathcal C}(\widehat{\mathbf V}_{\mathrm{GRPO}})\neq P_{\mathcal C}(V^\pi)$.
\end{theorem}

\noindent\textbf{Comment.}
Projecting per-prefix estimates into the tree-consistent class $\mathcal C$ cannot increase—and typically reduces—their distance to the best tree-consistent approximation of the truth.
Thus SAE is estimation-optimal \emph{within} the structural class encoded by the ordering constraints.

\vspace{1ex}
\noindent The appendix contains the formal assumptions, corollaries, and full proofs (Appendix~\ref{appendix:theory}).

\section{Experiment}

\subsection{Setup}

We train models on \textbf{GSM8K}~\citep{gsm8k}, a standard benchmark for multi-step mathematical reasoning, and evaluate them on GSM8K, \textbf{GSM-Symbolic}~\citep{gsm-symbolic} and \textbf{MATH}~\citep{math} to provide robust performance estimates and reduce reward exploitation.  For staged learning, we employ \texttt{GSM8K-MCTS}, an offline dataset generated via the MCTS-driven multi-turn chain-of-thought (CoT) procedure detailed in~\citep{rstar}. This dataset comprises tree-structured reasoning trajectories derived from 16 MCTS rollouts per problem, each with a maximum depth of 5 steps and up to 5 child nodes explored per step. These trajectories are \textit{augmented} with all their prefixes (e.g., ABCDE yields A, AB, ABC, ABCD), resulting in approx. 160k unique staged prefixes. Experiments employ \texttt{Qwen2.5-1.5B} as the student policy and \texttt{Qwen2.5-7B} as the teacher when applicable. The final answer accuracies (pass$@1$) on the test sets are reported.

\paragraph{Advantage estimation structures.}
We compare three structures for advantage estimation: \textit{flat} (vanilla per-step, no hierarchy), \textit{trace} (advantages along single rollouts without interleaving, used as ablation), and \textit{tree} (models full MCTS prefix hierarchy). 
\begin{table}[t]
\centering
\scriptsize
\begin{tabular}{llcccS[table-format=3.2]}
\toprule
\textbf{Method} & \textbf{Training Data} & \textbf{Advantage} & \boldmath$V(x)$ & \textbf{Acc. (\%)} & {\makecell[c]{\textbf{Constraint} \\ \textbf{Sat. (\%)}}} \\
\midrule
\rowcolor{gray!10}
GRPO~\citep{grpo} & GSM8K & Flat & -- & 76.27 & {--} \\
\midrule
\multirow{6}{*}{Tree-OPO} & \multirow{6}{*}{GSM8K-MCTS} & Flat & -- & 75.66 & {$79.26 \pm \text{\scriptsize 31.94}$} \\
\cellcolor{gray!10} & \cellcolor{gray!10} & \cellcolor{gray!10} Trace & \cellcolor{gray!10} -- & \cellcolor{gray!10} 73.91 & \cellcolor{gray!10} {--} \\
\cmidrule(lr){3-6}
& & \multirow{3}{*}{Tree (Heuristic)} & Expectation & \textbf{77.63} & {$97.98 \pm \text{\scriptsize 5.59}$} \\
\cellcolor{gray!10} & \cellcolor{gray!10} & & \cellcolor{gray!10} Optimistic & \cellcolor{gray!10} 70.58 & \cellcolor{gray!10} {--} \\
& & & Pessimistic & 67.40 & {--} \\
\cmidrule(lr){3-6}
\cellcolor{gray!10} & \cellcolor{gray!10} & \cellcolor{gray!10} SAE (\textit{Hard}) & \cellcolor{gray!10} -- & \cellcolor{gray!10} 75.21 & \cellcolor{gray!10} {$100 \pm \text{\scriptsize 0}$} \\
\cellcolor{gray!10} & \cellcolor{gray!10} & \cellcolor{gray!10} SAE (\textit{Soft}) & \cellcolor{gray!10} -- & \cellcolor{gray!10} 77.41 & \cellcolor{gray!10} {$100 \pm \text{\scriptsize 0}$} \\
\bottomrule
\end{tabular}
\caption{Comparison of GRPO and Tree-OPO variants on GSM8K. \textit{Constraint Sat.} measures the percentage of ordering constraints satisfied by raw rewards before applying SAE. A dash (--) denotes non-applicability or omission for auxiliary variants.}
\label{tab:result_Tree-OPO}
\end{table}

\subsection{Results}

\begin{figure*}[t]
  \centering
  \begin{subfigure}[t]{0.32\textwidth}
    \includegraphics[width=\linewidth]{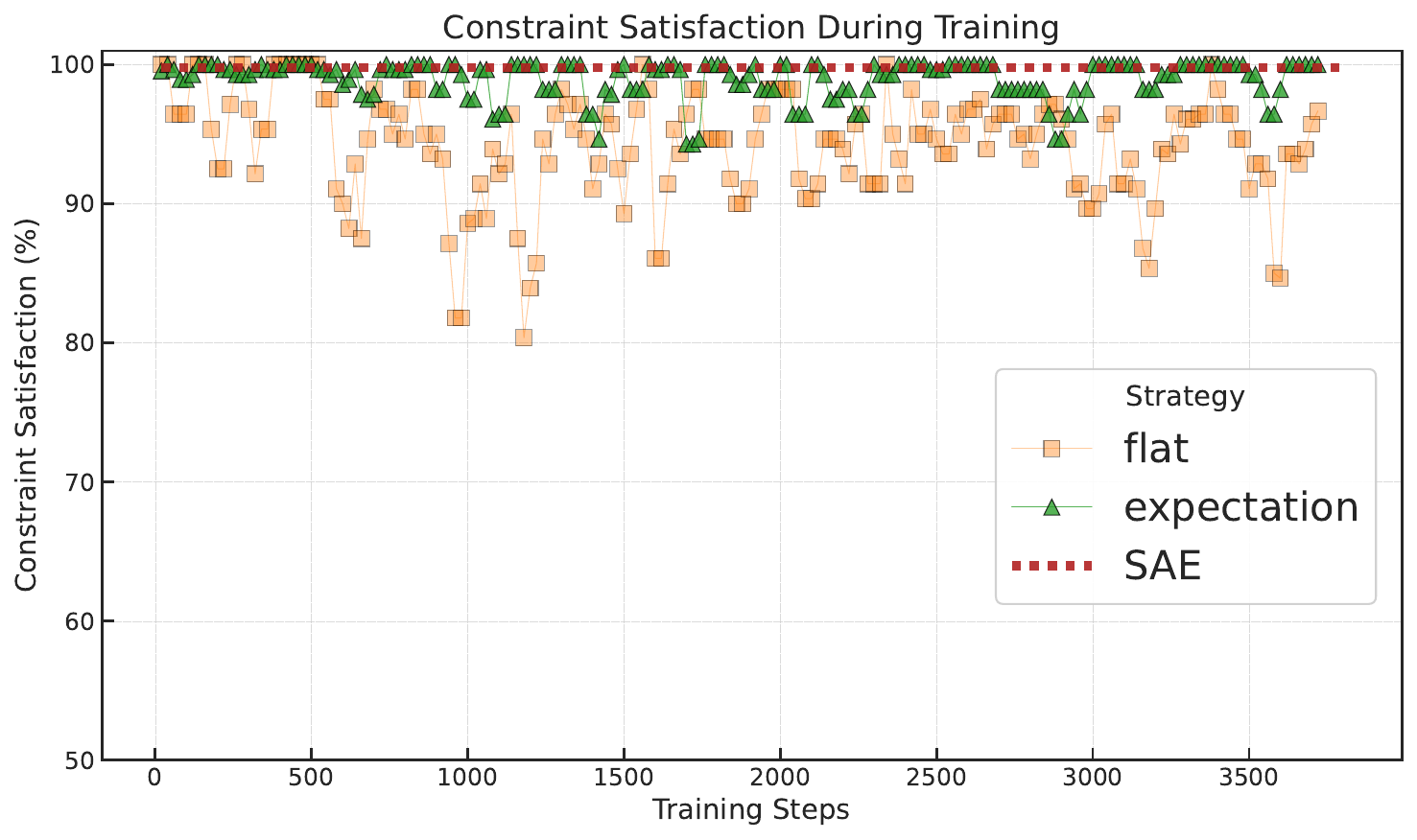}
    \caption{Constraint satisfaction}
    \label{fig:performance_a_constraint_sat}
  \end{subfigure}
  \hfill
  \begin{subfigure}[t]{0.32\textwidth}
    \includegraphics[width=\linewidth]{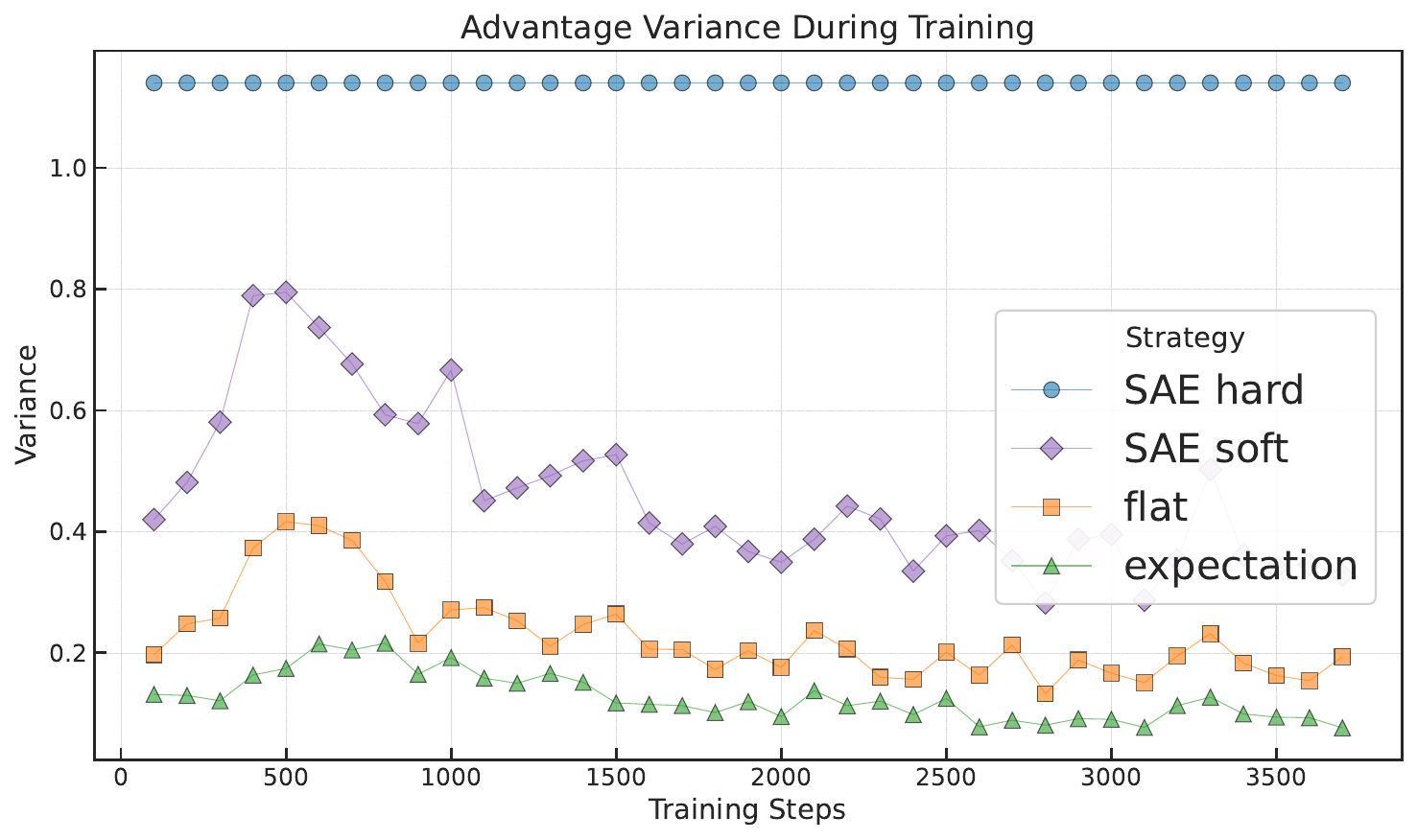}
    \caption{Advantage variance}
    \label{fig:performance_b_adv_var}
  \end{subfigure}
  \hfill
  \begin{subfigure}[t]{0.32\textwidth}
    \includegraphics[width=\linewidth]{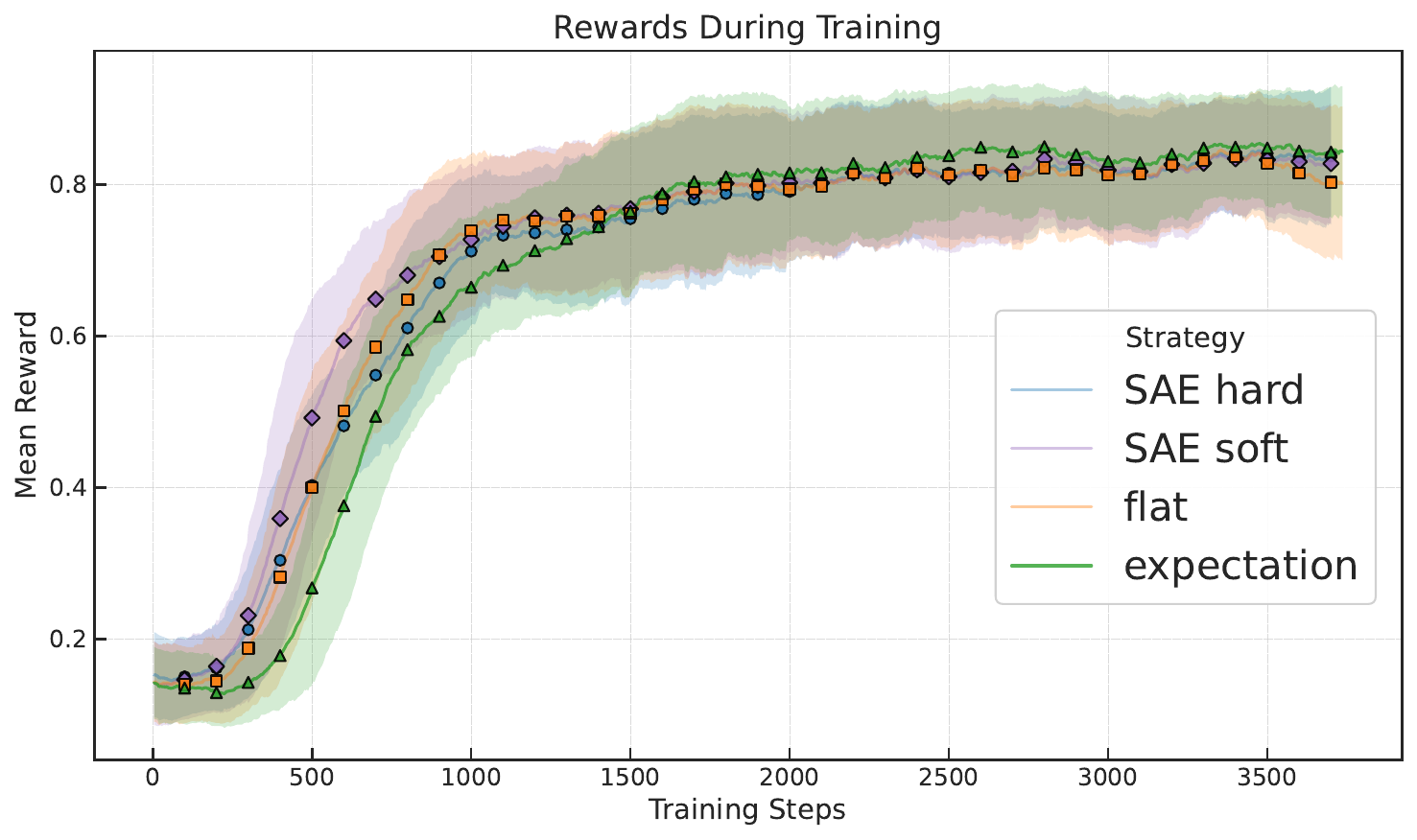}
    \caption{Group average reward}
    \label{fig:performance_c_reward}
  \end{subfigure}

  \vspace{-2mm}
  \caption{
    \textbf{Training metrics over time.} 
    (\subref{fig:performance_a_constraint_sat}) Satisfaction rate of prefix-ordering constraints imposed by structured advantage estimation.
    (\subref{fig:performance_b_adv_var}) Variance of advantage estimates; lower variance improves update stability.
    (\subref{fig:performance_c_reward}) Task-specific average rewards during training.
    All curves are smoothed; color indicates strategy.
  }
  \label{fig:overall_performance_metrics}
  \vspace{-3mm}
\end{figure*}

Tree-structured advantage estimation improves over both flat and trace-based baselines, achieving 77.63\% accuracy under the expectation heuristic. This setup best respects the MCTS-derived prefix hierarchy and yields the most informative gradient signal.
Among the baselines, the expectation-based \textbf{} \(V(x)\) outperforms both \textbf{optimistic} and \textbf{pessimistic} variants, suggesting that the empirical values best satisfy the ranking constraints and stabilize updates. 
While the gains over vanilla GRPO are modest, the comparison between Tree-OPO variants is more revealing. The expectation-based baseline achieves the highest accuracy, a result strongly correlated with its low advantage variance and nearly perfect tree-consistency constraint satisfaction (97.98\%, as shown in Figures~\ref{fig:performance_b_adv_var} and \ref{fig:performance_a_constraint_sat}).

In contrast, the \textbf{SAE (Hard)} variant serves as an ablative baseline testing the effect of enforcing perfect constraint satisfaction. Its non-convex normalization ($\|\bm a\|^2 = N$) fixes the advantage variance near~1.0, distorting the learning signal by decoupling it from the true reward scale. This rigidity leads to instability and degraded performance (\textbf{75.21\%}).  

Relaxing the constraint in the \textbf{SAE (Soft)} formulation ($\|\bm a\|^2 \le N$) yields a smoother balance between structural consistency and adaptive scaling, improving stability and performance to \textbf{77.41\%}. This places it on par with the best-performing \textbf{Expectation} heuristic (\textbf{77.63\%}), which benefits from its unbiased variance-minimizing nature (Lemma~\ref{lem:exp}). The convex SAE thus matches the leading heuristic while retaining theoretical guarantees of \emph{gradient} variance reduction (Thm.~\ref{theo:tree-signal}) and the interpretability of a structured projection. Empirically, the observed \emph{advantage} variance of SAE (Soft) can exceed that of Tree-OPO (Flat) at certain steps. This does not contradict the theory, which concerns the variance of the policy-gradient estimator under projections of the same reward vector. In practice, the soft formulation introduces relaxed penalties and numerical tolerances that can yield slightly higher apparent advantage variance when the norm constraint is not tightly enforced. Such deviations stem from implementation and tuning effects rather than a failure of the variance-reduction guarantee. Nevertheless, SAE consistently preserves the non-increasing variance behavior relative to its own projected inputs, validating the theoretical result in practice. The modest gains on GSM8K likely reflect the dataset’s simple tree structures, which limit the benefit of more sophisticated structural estimators.

\begin{wrapfigure}{r}{0.5\textwidth}
\vspace{-1em}
\centering
\includegraphics[width=0.48\textwidth]{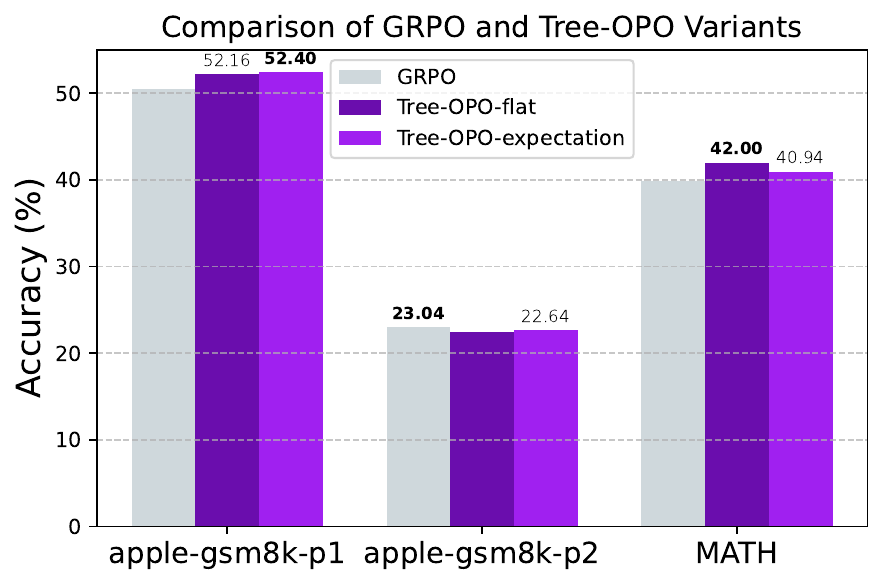}
\caption{Performance comparison of GRPO and Tree-OPO variants across different datasets.}
\label{fig:performance_barplot}
\vspace{-1em}
\end{wrapfigure}

\paragraph{Cross-Dataset Performance.} Figure~\ref{fig:performance_barplot} presents a comparison of the approaches' performance across different math-reasoning benchmarks. The results demonstrate the robustness of our methods while also showing that GRPO is a strong baseline, particularly when its global advantages naturally satisfy the tree-consistency constraints.

\subsection{Comparison of Policy Optimization Strategies}
\label{sec:efficiency_analysis}

We perform an ablation study comparing several policy optimization strategies for math reasoning tasks, focusing on how supervision signals, whether explicit distillation or task reward, affect final accuracy and resource usage. Table~\ref{tab:distillation_comparison} summarizes the reliance of each method on the teacher or critic models, the memory footprint, and the performance. Setup details are in Appendix~\ref{app:exp_setup}.

\begin{table}[H]
\centering
\scriptsize
\rowcolors{2}{gray!10}{white}
\begin{tabular}{lccccc}
\toprule
\textbf{Method} & \textbf{Teacher} & \textbf{Critic} & \textbf{Memory Cost} & \textbf{Acc. (\%)}\\
\midrule
REINFORCE KL~\citep{gkd} & \ding{51} & \ding{55} & High & 55.24 \\
REINFORCE KL + baseline~\citep{minillm} & \ding{51} & \ding{51} & High & 56.72 \\
Actor-Critic (TD-$\lambda$)~\citep{minillm} & \ding{51} & \ding{51} & High & 58.45 \\
\midrule
MCTS-driven REINFORCE & \ding{51} & \ding{55} & Medium & 67.55  \\
\midrule
Tree-OPO + REINFORCE KL & \ding{51} & (\ding{55}) & High & 69.90 \\
Tree-OPO (\textit{expectation}) & \ding{55} & \ding{55} & Low & \textbf{77.63} \\
\bottomrule
\end{tabular}
\caption{Ablation of policy optimization methods on GSM8K. Compared along supervision requirements, memory usage, and task accuracy.}
\label{tab:distillation_comparison}
\end{table}

\paragraph{KL-Distillation Methods.}
Methods such as \texttt{REINFORCE-based KL-distillation}~\citep{gkd} and \texttt{Actor-Critic methods}~\citep{minillm} supervise the student policy by minimizing (reverse) KL divergence to a teacher. While both distill teacher knowledge, they employ distinct underlying RL frameworks: the former utilizes the REINFORCE algorithm, whereas the latter adopts an Actor-Critic architecture with TD-$\lambda$ for value estimation. These approaches typically require online student/teacher's rollouts, teacher logits, and critic training (for Actor-Critic), resulting in high memory usage and limited gains (55–58\%). Our ablated method, \texttt{MCTS-driven REINFORCE}, uses MCTS trajectories to first fine-tune a teacher via supervised learning (SFT), and then applies \texttt{GKD}~\citep{gkd} (\texttt{REINFORCE} with reverse KL) to distill that teacher into the student. This avoids expensive teacher rollouts during student training, but still requires on-policy student rollouts, leading to improved efficiency compared to online teacher-guided methods.

\paragraph{Tree-OPO Variants.}
The final two rows of Table~\ref{tab:distillation_comparison} compare variants of our Tree-OPO. The \texttt{Tree-OPO + REINFORCE KL} baseline integrates both the task-specific reward and a KL divergence term from a fine-tuned teacher, introducing complexity in balancing multiple learning signals and increasing memory overhead, yet yielding only modest gains (69.90\%). Our key finding is that relying solely on task-specific rewards and advantage signals derived from MCTS-structured prefix trees, as in \texttt{Tree-OPO}, achieves better accuracy (77.63\%) with minimal supervision and resource costs.

\paragraph{Discussion.}
This ablation highlights the challenge of optimizing hybrid objectives in multi-task settings. While KL-based supervision provides stable gradients, it lacks explicit reward grounding and may underperform in generalization. Tree-OPO avoids this tradeoff by using structure-aware, reward-driven advantage estimation directly over staged reasoning prefixes—resulting in more efficient and effective policy learning.

\section{Conclusion}

We introduced \textbf{Tree-OPO}, a framework that enhances preference-based RL by replacing on-policy sampling with an off-policy curriculum of MCTS-derived prefixes. To address the staged setting this creates, we proposed \textbf{Staged Advantage Estimation (SAE)}, which computes low-variance, prefix-aware advantages by projecting rewards onto structural constraints. Leveraging the implicit \textbf{reverse curriculum} of the MCTS tree, Tree-OPO trains on a diverse mix of subproblems that stabilize and accelerate learning. We show theoretically that SAE reduces gradient variance and empirically that it improves both sample efficiency and accuracy over GRPO on mathematical reasoning tasks. Crucially, Tree-OPO is not merely another tree-based optimizer: by exploiting offline teacher prefixes it tackles the distinct challenge of heterogeneous prefix expectations, for which SAE provides a principled and effective solution. These findings open new directions for integrating structured off-policy data into policy optimization and for curriculum design in complex reasoning.


\medskip

\bibliographystyle{plainnat}
\bibliography{main}

\appendix
\include{appendix}

\end{document}

%% file: appendix.tex
\appendix
\input{proofs}
\subsection{On mean-centering without standardization}
\label{app:baseline-normalization}
A common practice in policy-gradient methods is to normalize advantages by both mean and standard deviation (z-scoring). However, in the staged setting this can distort reward magnitudes and interfere with the tree-structured consistency of prefix values. Our approach applies only mean-centering, which is both simpler and theoretically justified.

\begin{claim}[Mean-centering is the variance-optimal shift; scaling is conventional]
Let $\{A_k\}_{k=1}^K$ be advantage estimates with finite variance. Consider affine transforms $A'_k=(A_k-b)/c$ with $c>0$ and the stochastic policy-gradient estimator
\[
\widehat g \;=\; \frac{1}{K}\sum_{k=1}^K A'_k\,\nabla_\theta\log\pi_\theta(a_k\mid s_k).
\]
For any fixed $c>0$, the choice $b=\mathbb E[A]$ minimizes the (upper-bounded) variance of $\widehat g$. Moreover, changing $c$ simply rescales $\widehat g$ (and its variance) by $1/c$ (resp.\ $1/c^2$) and can be absorbed into the learning rate. Hence we set $c=1$ to preserve reward scale and avoid distorting tree-relative magnitudes.
\end{claim}

\begin{proof}
Using the standard bound on score-function moments (Assumption (A3)),
\[
\Var(\widehat g)
\;=\; \Var\!\Big(\frac{1}{K}\!\sum_{k=1}^K \tfrac{A_k-b}{c}\,\nabla_\theta\log\pi_\theta(a_k\mid s_k)\Big)
\;\le\; \frac{1}{K}\,\frac{G^2}{c^2}\,\Var(A-b).
\]
For fixed $c>0$, the right-hand side is minimized over $b$ by $b=\mathbb E[A]$, since $\Var(A-b)$ is minimized by mean-centering. The factor $1/c^2$ shows that $c$ only rescales the estimator and its variance; in practice $c$ trades off identically with the learning rate and does not improve the signal beyond a step-size change. We therefore adopt $c=1$ by convention to retain the native reward scale.
\end{proof}

This explains why simple mean-centering suffices: it provides the variance-optimal \emph{shift} while avoiding the instability and semantic distortion introduced by standardization of the \emph{scale}.\footnote{Z-scoring may aid cross-task comparability, but in tree-structured reasoning the absolute scale carries information about prefix difficulty and should be preserved.}

\subsection{Monte--Carlo baseline: unbiasedness and variance}
\label{app:mc_baseline}

Let $r^{(1)},\dots,r^{(M)}\in\{0,1\}$ be independent outcomes of rollouts from prefix $p$ under $\pi_\theta$.
Define the empirical estimator
\[
\widehat V_M(p)=\frac{1}{M}\sum_{m=1}^M r^{(m)}.
\]
Unbiasedness is immediate:
\[
\mathbb{E}[\widehat V_M(p)]=\mathbb{E}[r\mid p]=V(p).
\]
The variance satisfies
\[
\Var[\widehat V_M(p)]=\frac{\Var[r\mid p]}{M}=\frac{V(p)(1-V(p))}{M},
\]
since $r\in\{0,1\}$. By the Central Limit Theorem, for large $M$,
\[
\widehat V_M(p)\approx\mathcal{N}\!\big(V(p),\,V(p)(1-V(p))/M\big),
\]
so a (Gaussian) approximate \(1\!-\!\delta\) confidence interval has half-width \(\Phi^{-1}(1-\delta/2)\sqrt{V(p)(1-V(p))/M}\).
A conservative non-asymptotic bound follows from Hoeffding's inequality:
\[
\Pr\big(|\widehat V_M(p)-V(p)|\ge\epsilon\big)\le 2\exp(-2M\epsilon^2).
\]
These identities quantify the compute–variance tradeoff: sparse rewards (small \(V(p)\)) require large \(M\) to reduce baseline noise, motivating the low-cost, biased heuristics we use in practice (Sec.~\ref{sec:SAE}).

\section{Experimental Setup: Hyperparameters and Model Configuration}
\label{app:exp_setup}

This section details the hyperparameter and model settings used for both Tree-OPO and GPRO training as well as the training setting combines the distillation-based methods.

\paragraph{Policy Optimization Objectives.} Our policy $\pi_\theta$ is optimized using a combined objective, $L_{\text{policy}}$, which integrates a GRPO-style policy gradient term, \textit{optionally} with a distillation loss: $$L_{\text{policy}} = - \mathbb{E}_{\tau} \left[ \sum_{t=0}^{T-1} \log \pi_{\theta}(a_t|s_t) A^{\pi_{\theta}}(s_t, a_t) \right] + \beta_{\text{distill}} D_{\text{KL}}(\pi_{\text{teacher}}(a_t|s_t) \Vert \pi_{\theta}(a_t|s_t))$$ The first term represents the policy gradient loss (negative of the objective), where $A^{\pi_{\theta}}(s_t, a_t)$ is the advantage from Section~\ref{sec:SAE}. The $\alpha$ parameter (0.5) serves as the weighting coefficient for the baseline component within the advantage estimation ($a'_i = r_i - \alpha V(p_i)$). The second term is an optional distillation loss. When distillation is applied, it is weighed by the \texttt{distillation coefficient } ($\beta_{\text{distill}} = 0.1$). Otherwise, $\beta_{\text{distill}}$ is set to $0$. We utilize \textbf{reverse KL divergence}, $D_{\text{KL}}(P \Vert Q)$, where $P$ is the distribution of teacher policies ($\pi_{\text{teacher}}$) and $Q$ is the distribution of student policies ($\pi_{\theta}$). This choice promotes student coverage of the teacher's distributional modes, crucial for tasks with multiple valid reasoning paths. Separate learning rates are specified for the primary GRPO policy (\texttt{3e-5}) and for the distillation loss component (\texttt{5e-4}), or in specific distillation-only training regimes \textit{when distillation is enabled}.

\begin{table}[htbp]
\centering
\setlength{\abovecaptionskip}{10pt} 
\footnotesize
\begin{tabular}{lc}
\toprule
\textbf{Parameter} & \textbf{Value} \\
\midrule
$\alpha$ (Advantage Baseline Weight) & 0.5 \\
Distillation Coefficient ($\beta_{\text{distill}}$) & 0.1 \\
Model Precision & bf16 \\
Teacher Precision & bf16 \\
Group Size & 8 \\
Max Prompt Length & 256 (GRPO), 512 (Tree-OPO) \\
Max Sequence Length & 512 (GRPO), 768 (Tree-OPO) \\
Top-$k$ (Sampling) & 0 \\
Top-$p$ (Sampling) & 1.0 \\
Temperature (Sampling) & 1.0 \\
KL Divergence Type & Reverse \\
Optimizer & AdamW \\
Batch size & 32 \\
Learning rate (Policy, GRPO) & 3e-5 \\
Learning rate (Policy, Distillation) & 5e-4 \\
LR Scheduler & Cosine \\
Critic Head Architecture & Llama Decoder \\
Critic Warm-up Epochs & 20 \\
Critic Algorithm & Actor-Critic with TD($\lambda$) \\
GAE\_gamma ($\gamma$) & 0.95 \\
PEFT LoRA Rank (r) & 16 \\
PEFT LoRA Alpha ($\alpha_{\text{LoRA}}$) & 64 \\
PEFT LoRA Dropout & 0.1 \\
\bottomrule
\end{tabular}
\caption{Hyperparameter and model settings for Tree-PRO, encompassing GRPO and distillation components.}
\label{tab:hyperparam}
\end{table}

\paragraph{Critic Model and Loss.}
The critic network learns a state-value function, $V_\phi(s)$, to estimate expected returns from prefixes. Its \texttt{Llama Decoder} architecture allows it to process sequence contexts for value prediction \texttt{GAE\_gamma} ($\gamma = 0.95$) is the discount factor for Generalized Advantage Estimation (GAE). The critic's loss, $L_{\text{critic}}$, is a Mean Squared Error (MSE) between its prediction and the computed target value:
$$ L_{\text{critic}} = \frac{1}{N} \sum_{i=1}^N (V_\phi(s_i) - V^{\text{target}}_i)^2 $$
where $V^{\text{target}}_i$ is the Generalized Advantage Estimation (GAE) return for state $s_i$. 

\paragraph{Additional Model Settings.}
All models, including student and teacher, utilize \texttt{bf16} precision. Generation sampling employs \texttt{Top-p} ($p=1.0$) with \texttt{Temperature} $1.0$. The \texttt{Group Size} for GRPO batching is 8. Standard GRPO configurations use maximum prompt and sequence lengths of 256 and 512 tokens respectively, while Tree-OPO setups extend these to 512 and 768. Parameter Efficient Fine-Tuning (PEFT) via LoRA is applied, configured with a LoRA \texttt{rank} of 16, a scaling \texttt{alpha} of 64, and a \texttt{dropout} rate of 0.1.

\paragraph{SAE Formulations and Solver.}
We implement SAE by solving the constrained optimization problem in Eq.~\ref{eq:sae-opt}. We consider two primary formulations:

The \textbf{SAE (\textit{hard})} version, evaluated in our experiments, enforces a strict margin of $\delta=0.01$ on triplet-consistency constraints alongside the non-convex equality normalization $\|\bm a\|^2 = N$.

The \textbf{SAE (\textit{soft})} version is a convex relaxation used for our theoretical analysis. This formulation uses the inequality $\|\bm a\|^2 \le N$ and typically zero margins to guarantee a unique, well-behaved solution. An alternative soft approach converts the ordering constraints into a penalty term in the objective:
\[
\min_{\bm{a}\in\mathbb{R}^N} \quad \|\bm{a} - \bm{r}\|^2 + \lambda \sum_{(i,j) \in \mathcal{C}_{\text{order}}} \max(0, a_i - a_j + \delta)^2.
\]
Our current results are based on the \textit{hard}-constrained formulation; a full evaluation of the penalized soft version is deferred to future work. Both formulations can be solved using standard numerical methods; we use \texttt{scipy.optimize.minimize} with the SLSQP algorithm, warm-started from mean-centered rewards.

\section{Staged Advantage Estimation: A Minimal Example}
\label{appendix:sae}

To illustrate the structure of our staged prompting setup and the computation of advantages, we present a minimal example derived from the reasoning tree described in Figure~\ref{fig:MCTS_tree_curriculum}. Each prompt chain represents an incremental sequence of reasoning steps towards a solution, beginning with a base question (Q) and extended by alphabet-labeled stages (e.g., A, B, C).

\begin{figure}[h]
  \centering
  \begin{subfigure}[t]{0.22\linewidth}
    \centering
    \includegraphics[width=\linewidth]{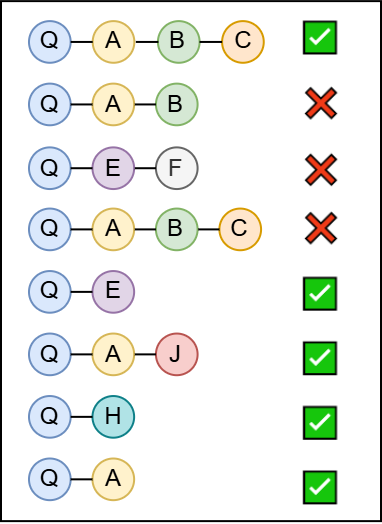}
    \caption{A group of staged prompts. }
    \label{fig:stage_prompts_demonstration}
  \end{subfigure}
  \hfill
  \begin{subfigure}[t]{0.72\linewidth}
    \centering
    \includegraphics[width=\linewidth]{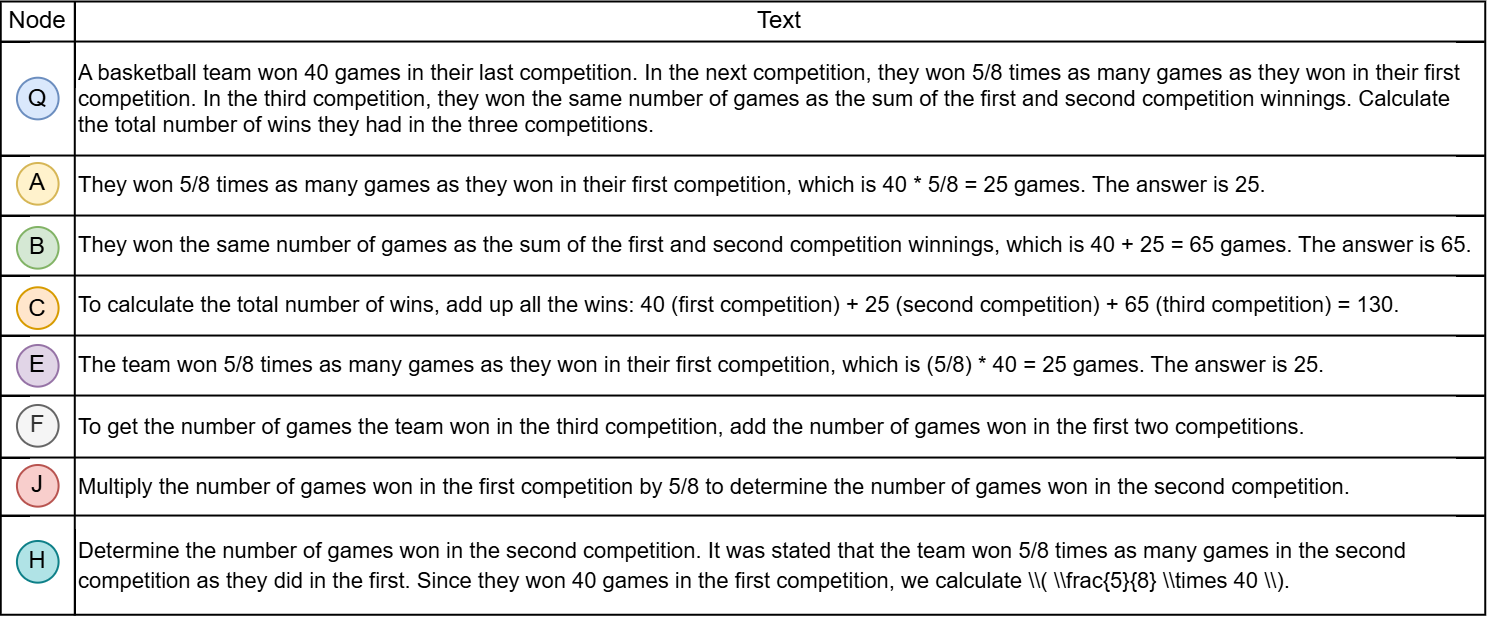}
    \caption{Correspondence between reasoning nodes and textual content. Each node represents a reasoning step in a multi-hop prompt.}
    \label{fig:node_text}
  \end{subfigure}
  \caption{\textbf{Illustration of Staged Reasoning Prompts and Text Mappings.} (a) A group of example staged prompts. Each row is a distinct prompt chain, and the checkmark/cross indicates whether the model produced a correct final answer when conditioned on that prompt. (b) Correspondence between symbolic reasoning nodes (Q, A, B, etc.) and their textual content. Each node represents a distinct reasoning step, forming compositional instructions separated by double newlines (\texttt{\char`\\n\char`\\n}) in the actual prompt fed to the language model.}
  \label{fig:staged_prompt_and_node_text}
\end{figure}

\noindent
Table~\ref{tab:potential_values} presents success statistics and heuristic value estimations ($V_{\text{exp}}(x)$, $V_{\text{opt}}(x)$, $V_{\text{pes}}(x)$). As detailed in Section \ref{sec:SAE}, advantage optimization is guided by \textbf{tree consistency constraints} ($\mathcal{C}_{\text{order}}$), ensuring coherent advantage signals.

The \textbf{expected value baseline} ($V_{\text{exp}}(x)$), calculated as average success rate, inherently aligns with these constraints, yielding precise 'surprise' signals. For instance, `$\mathcal{C}_{\text{pair}}$` requires $a_i < a_j$ for a failing parent $p_i$ leading to a successful child $p_j$. Comparing `Q-A-B` ($E[R]=1/3$, $r=0$) and its successful descendant `Q-A-B-C` ($E[R]=1/2$, $r=1$ from `t0`), $V_{\text{exp}}$ generates advantages of $-1/6$ and $3/4$ respectively, satisfying $a_i < a_j$. While this example doesn't fully capture all `$\mathcal{C}_{\text{triplet}}$` conditions, $V_{\text{exp}}$'s accurate value estimations ($E[R|\text{Q-A-B}]=1/3$ vs $E[R|\text{Q-E-F}]=0$) are fundamental for enabling such fine-grained control over advantage ordering.

Conversely, \textbf{optimistic} ($V_{\text{opt}}(x)$) and \textbf{pessimistic} ($V_{\text{pes}}(x)$) baselines often violate these constraints. $V_{\text{pes}}(\text{Q-E-F})=0$ and $V_{\text{pes}}(\text{Q-A-B})=0$ despite different $E[R]$. Similarly, $V_{\text{opt}}$ assigns 1 to `Q-A-B`, `Q-A-B-C`, and `Q-A-J` (all having different $E[R]$), losing vital discriminative power for tree consistency.

\begin{table}[h]
\centering
\small
\renewcommand{\arraystretch}{1.2}
\begin{tabular}{|>{\centering\arraybackslash}m{3cm}|c|c|c|c|c|}
\hline
\textbf{x / Prompt} & \textbf{Success} & \textbf{Total} & \boldmath$V_{\text{exp}}(x)$ & \boldmath$V_{\text{opt}}(x)$ & \boldmath$V_{\text{pes}}(x)$ \\
\hline
Q-A       & 3 & 5 & $\frac{3}{5}$ & 1 & 0 \\
Q-A-B     & 1 & 3 & $\frac{1}{3}$ & 1 & 0 \\
Q-A-B-C   & 1 & 2 & $\frac{1}{2}$ & 1 & 0 \\
Q-A-J     & 1 & 1 & 1             & 1 & 0 \\
Q-E       & 1 & 2 & $\frac{1}{2}$ & 1 & 0 \\
Q-E-F     & 0 & 1 & 0             & 0 & 0 \\
Q-H       & 1 & 1 & 1             & 1 & 1 \\
\hline
\end{tabular}
\caption{Success statistics and corresponding heuristic value baselines for example reasoning prefixes. $V_{\text{exp}}(x)$, $V_{\text{opt}}(x)$, and $V_{\text{pes}}(x)$ are derived from observed success rates within the prefix's subtree, reflecting different assumptions of future success.}
\label{tab:potential_values}
\end{table}

\noindent
\noindent
Table~\ref{tab:final_advantage} demonstrates staged advantage computation. The \textbf{expectation baseline} ($V_{\text{exp}}$) best satisfies the \emph{tree consistency constraints}. Its accurate reflection of expected rewards ensures advantages consistently signal better/worse-than-expected outcomes relative to a path's true potential. This yields informative, context-sensitive advantages. In contrast, $V_{\text{opt}}$ and $V_{\text{pes}}$'s coarse assignments lead to misleading signals, failing tree consistency.

\begin{table}[htbp]
\centering
\small
\renewcommand{\arraystretch}{1.3}
\begin{tabular}{|c|>{\columncolor{green!15}}c|c|c|}
\hline
\textbf{R} & \boldmath$R - \frac{1}{2} V_{\text{exp}}(x)$ & \boldmath$R - \frac{1}{2} V_{\text{opt}}(x)$ & \boldmath$R - \frac{1}{2} V_{\text{pes}}(x)$ \\
\hline
1 & $\frac{7}{10}$ & $\frac{1}{2}$ & 1 \\
0 & $-\frac{1}{6}$ & $-\frac{1}{2}$ & 0 \\
0 & 0 & $-\frac{1}{2}$ & 0 \\
0 & $-\frac{1}{4}$ & $-\frac{1}{2}$ & 0 \\
1 & $\frac{3}{4}$ & $\frac{1}{2}$ & 1 \\
1 & $\frac{1}{2}$ & $\frac{1}{2}$ & 1 \\
1 & $\frac{1}{2}$ & 1 & 1 \\
1 & $\frac{3}{5}$ & $\frac{1}{2}$ & $\frac{1}{2}$ \\
\hline
\end{tabular}
\caption{Example staged advantage values computed as $R - \alpha V(x)$ for different heuristic baselines. The \textbf{expectation baseline} ($V_{\text{exp}}$; second column, shaded) better satisfies the \emph{tree-consistency constraints}, reflecting coherent advantage values aligned with reward structure across the tree. In contrast, the \textbf{optimistic} ($V_{\text{opt}}$) and \textbf{pessimistic} ($V_{\text{pes}}$) baselines exhibit more inconsistent or overly conservative adjustments. Each row corresponds to a prompt-completion trajectory with binary reward $R$ from Figure~\ref{fig:staged_prompt_and_node_text}(a).}
\label{tab:final_advantage}
\end{table}

\section{Multi-Turn MCTS Rollout Structure and Tree Visualization}
\label{app:teacher_prefixes}
\begin{sidewaysfigure}
    \centering
    \includegraphics[width=\textheight]{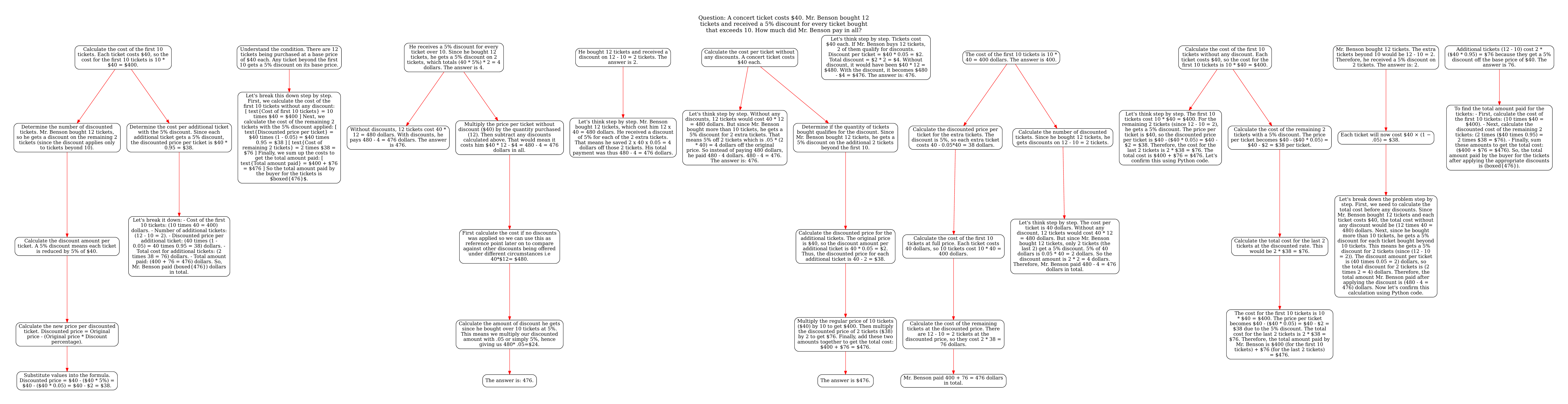}
    \caption{\centering A visualization of a multi-turn MCTS tree to a question in GSM8k.}
    \label{fig:mcts-tree}
\end{sidewaysfigure}

\paragraph{Multi-turn Rollout.}  
Following rStar~\cite{rstar}, a reasoning episode is decomposed into stages \(t = 1, \dots, T\), where at each stage the agent generates an action \(a_t\) conditioned on a state \(s_t\), and transitions to the next state via \(s_{t+1} = s_t \circ a_t\). This defines a multi-turn rollout, where the final solution is constructed as a chain of reasoning steps \(\{a_1, \dots, a_T\}\). Unlike standard single-step MCTS, which performs search from a fixed root to a terminal state, rStar applies MCTS recursively over multiple stages, building up partial completions and branching decisions at each intermediate node. The result is a structured, non-Markovian tree of reasoning trajectories, where each visited prefix \(s_t\) corresponds to a meaningful subproblem along a candidate path. Thus, the training data is not drawn directly from an offline teacher policy, but rather sampled from a \emph{complex, multi-turn distribution} induced by search—reflecting the teacher’s internal planning dynamics across many possible intermediate reasoning paths. Figure~\ref{fig:mcts-tree} iillustrates a reasoning tree produced by the multi-turn MCTS rollout process.

\paragraph{Learning from the MCTS-Induced Distribution.}
Although the MCTS-derived distribution over prefixes does not correspond to samples from a stationary expert policy, it is shaped by a structured search process guided by value estimates and exploration. Each prefix \(s_t\) encountered during multi-turn rollout reflects a plausible intermediate reasoning state prioritized during search. From an RL perspective, this yields an offline dataset composed of staged pairs \((s_t, c_t, R_t)\), where completions \(c_t\) and rewards \(R_t\) are conditioned on prefix context. This structured distribution captures diverse yet high-quality trajectories that extend beyond single-path demonstrations. By leveraging this data with policy gradient methods and advantage estimation, the student can effectively optimize its policy without requiring direct imitation. Thus, the learning process exploits the inductive bias of the MCTS-induced distribution to supervise training in a way that aligns with structured reasoning dynamics.

\newpage

\section{Algorithm}

\begin{algorithm}[h]
\DontPrintSemicolon
\caption{Tree-OPO: Tree-structured Relative Policy Optimization}
\label{alg:Tree-OPO_compact}
\KwIn{Teacher MCTS budget $(B, d_{\max}, b)$; student policy $\pi_\theta$; batch size $K$; baseline rollouts $M$; scaling $\alpha$; solver option; learning rate $\eta$}
\KwOut{Updated parameters $\theta$}

\textbf{Offline MCTS:}\;
\ForEach{training problem}{
  Run multi-turn MCTS ($B$ rollouts, depth $\le d_{\max}$, branching $\le b$)\;
  Collect trajectories and augment with all prefixes $\rightarrow$ prefix tree $\mathcal T$\;
}

\While{not converged}{
  Sample $K$ prefixes $\{p_k\}\subset\mathcal T$\;
  \For{$k=1,\dots,K$}{
    Rollout $\hat c_k\sim\pi_\theta(\cdot\mid p_k)$; reward $r_k\in\{0,1\}$\;
  }
  Estimate baseline $V(p_k)$ (\emph{MC}, empirical, optimistic/pessimistic)\;
  Compute $a'_k \gets r_k - \alpha V(p_k)$; mean-center $\{a'_k\}\to\{a_k\}$\;
  \If{SAE enabled}{
    Refine $\{a_k\}$ via heuristic or constrained QP\;
  }
  $\widehat g \gets \tfrac{1}{K}\sum_k a_k\nabla_\theta\log\pi_\theta(\hat c_k\mid p_k)$\;
  Update $\theta \gets \theta + \eta\,\widehat g$\;
}
\end{algorithm}

%% file: proofs.tex
\section{Theoretical Proofs}
\label{appendix:theory}

\paragraph{Standing assumptions.}
We make the following assumptions throughout the appendix (stated explicitly here to keep proofs self-contained):
\begin{enumerate}[label=(A\arabic*)]
  \item \textbf{Consistent ordering.} The ordering constraint set $\mathcal C_{\mathrm{order}}$ is acyclic (i.e., contains no contradictory pairs).
  \item \textbf{Convexified normalization.} For theoretical statements we replace the RMS equality by the convex ball constraint $\|\mathbf a\|^2 \le N$ (practical solvers may enforce equality or use penalties; see discussion in the main text). This yields a closed, convex and bounded feasible set when combined with linear constraints.
  \item \textbf{Bounded policy gradients.} The policy is such that $\mathbb{E}_{p,a}\bigl[\|\nabla_\theta\log\pi_\theta(a\mid p)\|^2\bigr] \le G^2$.
\end{enumerate}

\subsection{Proof of Lemma~\ref{lem:unbias}: Unbiasedness of Gradient Estimate}
\label{app-lem:unbias}

\begin{lemma}[Unbiasedness of Gradient Estimate~\cite{williams1992simple}]
Let $a\sim\pi_\theta(\cdot\mid p)$ and let $V(p)$ be any deterministic function of the prefix $p$. Define
\[
\widehat g=(r-\alpha V(p))\,\nabla_\theta\log\pi_\theta(a\mid p).
\]
Then $\mathbb{E}_{p,a}[\widehat g]=\nabla J(\theta)$.
\end{lemma}

\begin{proof}
By the policy gradient theorem with a state-dependent baseline $b(p)$:
\[
\nabla J(\theta)=\mathbb{E}_{p,a}\big[(r-b(p))\nabla_\theta\log\pi_\theta(a\mid p)\big].
\]
Set $b(p)=\alpha V(p)$. Then
\[
\mathbb{E}_{p,a}[\alpha V(p)\nabla_\theta\log\pi_\theta(a\mid p)]
=\mathbb{E}_p\!\Big[\alpha V(p)\,\mathbb{E}_{a\sim\pi_\theta(\cdot\mid p)}[\nabla_\theta\log\pi_\theta(a\mid p)]\Big].
\]
But $\mathbb{E}_{a\sim\pi}[\nabla_\theta\log\pi_\theta(a\mid p)]=\nabla_\theta\sum_a\pi_\theta(a\mid p)=\nabla_\theta 1=0$, so the baseline term vanishes. Hence $\mathbb{E}[\widehat g]=\nabla J(\theta)$.
\end{proof}

\subsection{Proof of Lemma~\ref{lem:exp}: Optimality of Expectation Baseline}
\label{app-lem:exp}

\begin{lemma}[Expectation Minimizes Variance\cite{greensmith2004variance}]
Among all deterministic functions $V(p)$, the choice $V(p)=\mathbb{E}[r\mid p]$ (and $\alpha=1$) minimizes $\Var[r-\alpha V(p)]$.
\end{lemma}

\begin{proof}
Using the law of total variance for $A=r-\alpha V(p)$:
\[
\Var[A]=\mathbb{E}_p[\Var[A\mid p]]+\Var_p[\mathbb{E}[A\mid p]].
\]
Since $V(p)$ is deterministic given $p$, $\Var[A\mid p]=\Var[r\mid p]$, independent of $V$. Moreover
\[
\mathbb{E}[A\mid p]=\mathbb{E}[r\mid p]-\alpha V(p).
\]
Thus
\[
\Var[A]=\mathbb{E}_p[\Var[r\mid p]]+\Var_p[\mathbb{E}[r\mid p]-\alpha V(p)].
\]
The first term is fixed; the second is minimized by setting $\alpha V(p)=\mathbb{E}[r\mid p]$, which yields the stated result.
\end{proof}

\subsection{Proof of Lemma~\ref{lem:tree-structure}: Tree-Induced Advantage Structure}
\label{app-lem:tree}

We now formalize the SAE projection used in the main text in a convex form suitable for theoretical guarantees.

\begin{lemma}[Tree-Induced Advantage Structure]
Let $\bm r\in\mathbb R^N$ be the observed rewards in a group and let $\mathcal C_{\mathrm{order}}$ be an acyclic set of pairwise ordering relations. Consider the convex QP
\begin{equation}
\begin{aligned}
\bm a^* &= \arg\min_{\bm a\in\mathbb R^N}\ \|\bm a-\bm r\|^2 \\
\text{s.t.}\quad & \mathbf 1^\top\bm a = 0,\qquad \|\bm a\|^2 \le N,\\
& a_i + \delta_{ij} \le a_j \quad \forall (i,j)\in\mathcal C_{\mathrm{order}},
\end{aligned}
\tag{\ref{eq:sae-opt}--convex}
\end{equation}
with fixed margin $\delta_{ij}\ge 0$. Under Assumption (A\!1) the feasible set is nonempty and convex, the objective is strictly convex, and the problem has a unique solution $\bm a^*$. Moreover $\bm a^*$ satisfies all linear ordering constraints, so that for each $(i,j)\in\mathcal C_{\mathrm{order}}$ we have $a_i^*+\delta\le a_j^*$.
\end{lemma}

\begin{proof}
With linear equalities/inequalities and the convex ball $\|\bm a\|^2\le N$, the feasible set is convex and closed. The objective $\|\bm a-\bm r\|^2$ is strictly convex (Hessian $2I\succ 0$). A strictly convex function on a nonempty convex compact set has a unique minimizer, so $\bm a^*$ exists and is unique. The ordering constraints are explicit linear constraints in the program, hence the minimizer satisfies them by construction.
\end{proof}

\paragraph{Practical remark.}
In practice we solve a softened variant (penalty or SLSQP with equality RMS); Appendix~\ref{app:exp_setup} discusses numerical choices. The convex formulation above is used only for the theoretical guarantees (existence, uniqueness, and structural encoding).

\subsection{Proof of Theorem~\ref{theo:tree-signal}: Gradient Signal Improvement}
\label{app-theo:tree-signal}

\begin{theorem}[Gradient Signal and Structure]
\label{app-theo:gradient-signal}
Under assumptions (A1)--(A3), let $\bm r\in\mathbb R^N$ be observed rewards in a group. Define the centered reward vector
\[
\bm r_0 \;=\; \bm r - \overline{\bm r}\,\mathbf 1,
\]
and let
\[
\mathcal F_0 \;=\; \{\,\bm a\in\mathbb R^N:\; \mathbf 1^\top\bm a=0,\; \|\bm a\|^2\le N,\; L\bm a\le \mathbf 0\,\},
\]
where the rows of $L$ are $e_i^\top-e_j^\top$ for each $(i,j)\in\mathcal C_{\mathrm{order}}$ (zero-margin pairwise orderings). Assume $\mathbf0\in\mathcal F_0$ (holds under zero margins and acyclicity). Let
\[
\bm a^* \;=\; \arg\min_{\bm a\in\mathcal F_0} \|\bm a - \bm r_0\|^2
\]
be the Euclidean projection of $\bm r_0$ onto $\mathcal F_0$. Then:
\begin{enumerate}[nosep]
  \item \textbf{(Variance non-increase)} The projection does not increase variance:
  \[
  \Var[\bm a^*] \;\le\; \Var[\bm r_0] \;=\; \Var[\bm r],
  \]
  where $\Var[\bm x] = \tfrac{1}{N}\|\bm x\|^2$ for zero-mean vectors.
\end{enumerate}
Moreover, equality holds if and only if $\bm r_0 \in \mathcal{F}_0$; otherwise the inequality is strict.
\end{theorem}

\begin{proof}

\noindent\textbf{Variance non-increase via convex projection.}
By construction $\bm a^*$ is the Euclidean projection of $\bm r_0$ onto the closed convex set $\mathcal F_0$. Projection optimality (firm nonexpansiveness / characterization of the projector; see \cite[Thm.~3.14]{bauschke2017convex}) gives the variational inequality
\[
\langle \bm r_0 - \bm a^*,\, \bm z - \bm a^* \rangle \le 0
\qquad\forall \bm z\in\mathcal F_0.
\]
Since $\mathbf0\in\mathcal F_0$ we may take $\bm z=\mathbf0$, yielding
\[
\langle \bm r_0 - \bm a^*,\ -\bm a^* \rangle \le 0
\quad\Longrightarrow\quad
\langle \bm r_0 - \bm a^*,\ \bm a^* \rangle \ge 0.
\]
Hence
\[
\|\bm a^*\|^2 \le \langle \bm r_0,\bm a^* \rangle \le \|\bm r_0\|\,\|\bm a^*\|,
\]
and dividing by $\|\bm a^*\|$ (if $\bm a^*\neq\mathbf0$) gives $\|\bm a^*\|\le\|\bm r_0\|$. If $\bm a^*=\mathbf0$ the inequality is trivial. Because both $\bm a^*$ and $\bm r_0$ are zero-mean, their population variances satisfy
\[
\Var[\bm a^*] \;=\; \frac{1}{N}\|\bm a^*\|^2 \;\le\; \frac{1}{N}\|\bm r_0\|^2 \;=\; \Var[\bm r_0].
\]
This establishes the variance non-increase property.

The inequality is strict exactly when $\bm r_0\notin\mathcal F_0$ (otherwise the projection returns $\bm r_0$ and norms are equal). Thus whenever the centered rewards violate at least one tree constraint the projection strictly reduces squared norm (hence variance).

\noindent\textbf{Role of the ordering constraints.}
The linear order constraints $L \bm a \le 0$ encode \emph{tree consistency} by enforcing prefix-relative advantage orderings. 
With the convexified normalization $\|\bm a\|^2 \le N$, the feasible set
\[
\mathcal F_0 = \bigl\{ \bm a : \mathbf 1^\top \bm a = 0,\;\; \|\bm a\|^2 \le N,\;\; L \bm a \le 0 \bigr\}
\]
is closed, convex, and contains $\mathbf 0$. Therefore, the Euclidean projection
\[
\bm a^* = \mathrm{Proj}_{\mathcal F_0}(\bm r_0)
\]
satisfies $\|\bm a^*\| \le \|\bm r_0\|$ and hence $\Var[\bm a^*] \le \Var[\bm r]$.  

If the unconstrained (GRPO) projection already respects the orderings, the two projections coincide. Otherwise, $\bm a^*$ enforces the tree structure by making the minimal modification (in Euclidean norm) to $\bm r_0$ necessary to satisfy the order constraints. In practice, this structural correction typically—and often strictly—reduces the advantage variance across prefixes when $\bm r_0$ violates tree constraints. 

Note, however, that there is no universal monotonicity guarantee comparing 
$\|\mathrm{Proj}_{\mathcal F_0}(\bm r_0)\|$ with $\|\mathrm{Proj}_{H}(\bm r_0)\|$ for every possible $\bm r_0$ (projections onto nested convex sets need not be ordered). Consequently, we present the projection variance bound as the rigorous core result and treat the additional empirical benefit relative to plain GRPO as an expected, often-observed effect supported by experiments.

\end{proof}

\subsection{GRPO as a Degenerate Projection and Variance Comparison}
\label{app-cor:grpo}

\begin{corollary}[SAE does not increase variance; strict reduction under violations]
\label{cor:grpo-compact}
Let $\mathbf r\in\mathbb R^N$ be observed rewards and $\mathbf r_0=\mathbf r-\bar r\mathbf1$ the centered rewards.
Assume $\mathcal F\subseteq H:=\{\mathbf a:\mathbf1^\top\mathbf a=0\}$ is closed, convex and satisfies $\mathbf0\in\mathcal F$ and $\|\mathbf a\|^2\le N$ for all $\mathbf a\in\mathcal F$. Denote
\(\mathbf a^*=\Pi_{\mathcal F}(\mathbf r_0)\) (SAE projection) and
\(\mathbf a_{\mathrm{GRPO}}=\mathbf r_0\) (plain mean-centering).
If $\mathbf r_0\neq\mathbf0$ define the std-normalized GRPO vector
\(\mathbf a_{\mathrm{GRPO\_std}}=\sqrt{N}\,\mathbf r_0/\|\mathbf r_0\|\).
Then:
\begin{enumerate}[nosep]
  \item \(\Var[\mathbf a^*]\le \Var[\mathbf r_0]=\Var[\mathbf a_{\mathrm{GRPO}}]\), with strict inequality whenever \(\mathbf r_0\notin\mathcal F\).
  \item For the nondegenerate case \(\mathbf r_0\neq\mathbf0\), \(\Var[\mathbf a^*]\le 1=\Var[\mathbf a_{\mathrm{GRPO\_std}}]\); hence SAE never increases variance relative to GRPO with std-normalization.
\end{enumerate}
In particular, when centered rewards violate tree-order constraints the SAE projection typically (and often strictly) reduces variance compared to plain mean-centering.
\end{corollary}

\begin{proof}
(1) By definition $\mathbf a^*$ is the Euclidean projection of $\mathbf r_0$ onto the closed convex set $\mathcal F$. Projection optimality yields, for all $\mathbf z\in\mathcal F$,
\(\langle \mathbf r_0-\mathbf a^*,\ \mathbf z-\mathbf a^*\rangle\le0\). Taking $\mathbf z=\mathbf0\in\mathcal F$ gives \(\langle \mathbf r_0-\mathbf a^*,\ \mathbf a^*\rangle\ge0\). Expanding norms:
\[
\|\mathbf r_0\|^2 = \|\mathbf r_0-\mathbf a^*\|^2 + 2\langle \mathbf r_0-\mathbf a^*,\mathbf a^*\rangle + \|\mathbf a^*\|^2
\ge \|\mathbf a^*\|^2,
\]
hence \(\|\mathbf a^*\|^2\le\|\mathbf r_0\|^2\). Dividing by \(N\) gives \(\Var[\mathbf a^*]\le\Var[\mathbf r_0]\). If \(\mathbf r_0\notin\mathcal F\) then \(\|\mathbf r_0-\mathbf a^*\|>0\) and the inequality is strict.

(2) For \(\mathbf r_0\neq\mathbf0\) the std-normalized GRPO vector has \(\|\mathbf a_{\mathrm{GRPO\_std}}\|^2=N\), so \(\Var[\mathbf a_{\mathrm{GRPO\_std}}]=1\). Since by assumption every \(\mathbf a\in\mathcal F\) satisfies \(\|\mathbf a\|^2\le N\), we have \(\|\mathbf a^*\|^2\le N\), hence \(\Var[\mathbf a^*]\le1=\Var[\mathbf a_{\mathrm{GRPO\_std}}]\). This proves that SAE does not increase variance compared to GRPO with std-normalization.
\end{proof}



\subsection{SAE Improves Estimation of the Best Consistent Value Function}
\label{sec:sae_closeness}

Recall from Lemma~\ref{lem:exp} that the variance-minimizing deterministic baseline for a prefix is the true prefix-value
\[
V^\pi(p) \;=\; \mathbb{E}_{\tau\sim\pi}\bigl[r(\tau)\mid \tau\supset p\bigr],
\]
i.e.\ the expected terminal reward of trajectories that extend prefix \(p\).

We consider three concrete empirical/algorithmic estimators of \(V^\pi\):

\begin{align}
\textbf{(GRPO, global)}\qquad
V_{\mathrm{GRPO}} &\;=\; \bar r \;=\; \frac{1}{N}\sum_{i=1}^N r(\tau_i), \label{eq:V_grpo}\\[4pt]
\textbf{(Subtree empirical / \(V_E\))}\qquad
V_{E}(p) &\;=\; \frac{1}{N_p^{\mathrm{sub}}}\sum_{\tau_i : \tau_i\supset p} r(\tau_i),
\label{eq:V_E}\\[4pt]
\textbf{(SAE / projected)}\qquad
\widehat{\mathbf V}_{\mathrm{SAE}} &\;=\; P_{\mathcal C}\!\bigl(\widehat{\mathbf V}_{\mathrm{GRPO}}\bigr),
\qquad
\widehat V_{\mathrm{SAE}}(p) \;=\; \bigl(\widehat{\mathbf V}_{\mathrm{SAE}}\bigr)_p.
\label{eq:V_SAE_prefix}
\end{align}

Here \(\{\tau_i\}_{i=1}^N\) are the sampled completions, \(N_p^{\mathrm{sub}}\) counts all sampled completions that extend prefix \(p\), and \(\widehat{\mathbf V}_{\mathrm{GRPO}}\in\mathbb R^P\) is the vector of per-prefix empirical exact-start means
\(\widehat V_{\mathrm{GRPO}}(p)=\tfrac{1}{n_p^{\mathrm{exact}}}\sum_{i:s_i=p} r(\tau_i)\) (used as the input to projection).
The set \(\mathcal C\subset\mathbb R^P\) denotes the closed convex class of prefix-value vectors satisfying the tree-ordering constraints (convexified as needed).

Equation~\eqref{eq:V_SAE_prefix} is the \emph{idealized} prefix-space SAE: it projects the empirical per-prefix vector onto \(\mathcal C\). In practice we realize SAE by a sample-space projection (solve the QP in sample-space for advantages, then aggregate), which is equivalent to the prefix-space projection under Assumption~A4 (constraints act only on prefix-aggregates); otherwise the two may differ slightly due to solver/regularization details.

\begin{theorem}[SAE optimally reduces estimation-to-class error]
\label{thm:sae-prefix-proj}
Let \(\mathbf V^*_{\mathcal C}=P_{\mathcal C}(\mathbf V^\pi)\) be the projection of the true prefix-value vector onto \(\mathcal C\). Then for any empirical GRPO prefix-vector \(\widehat{\mathbf V}_{\mathrm{GRPO}}\),
\[
\big\| \widehat{\mathbf V}_{\mathrm{SAE}} - \mathbf V^*_{\mathcal C} \big\|_2
\;\le\;
\big\| \widehat{\mathbf V}_{\mathrm{GRPO}} - \mathbf V^*_{\mathcal C} \big\|_2,
\]
with strict inequality whenever $\widehat{\mathbf V}_{\mathrm{GRPO}}\notin\mathcal C$ \emph{and} $P_{\mathcal C}(\widehat{\mathbf V}_{\mathrm{GRPO}})\neq P_{\mathcal C}(\mathbf V^\pi)$.
\end{theorem}

\begin{proof}[Sketch]
Immediate from the distance-decreasing property of Euclidean projection onto a closed convex set: for every \(z\in\mathcal C\) and any \(x\in\mathbb R^P\) we have \(\|P_{\mathcal C}(x)-z\|_2\le\|x-z\|_2\). Set \(x=\widehat{\mathbf V}_{\mathrm{GRPO}}\) and \(z=\mathbf V^*_{\mathcal C}\).
\end{proof}

\paragraph{Comparison and limitations.}
\begin{itemize}[nosep]
\item \textbf{\(V_E\) vs \(V_{\mathrm{GRPO}}\).} The subtree estimator \(V_E(p)\) (Eq.~\ref{eq:V_E}) trades variance for potential bias at each prefix. The MSE of an estimator is decomposed into its squared bias and variance: $\mathrm{MSE}(\widehat{V})=\mathrm{Bias}(\widehat{V})^2+\mathrm{Var}(\widehat{V})$. While \(V_{\mathrm{GRPO}}\) is an unbiased estimator of \(V^\pi(p)\) (zero bias, high variance), \(V_E(p)\) is a biased estimator (non-zero bias, low variance) due to pooling samples from different states. A good \(V_E\) estimator achieves a larger reduction in variance than the square of its introduced bias.
\item \textbf{\(V_{\mathrm{SAE}}\) vs \(V_{\mathrm{GRPO}}\).} Theorem~\ref{thm:sae-prefix-proj} guarantees that projecting GRPO estimates into \(\mathcal C\) reduces the estimation-to-class distance in \(\ell_2\). This is an {\em estimation} improvement relative to the feasible class \(\mathcal C\), but it does not by itself imply \(\widehat V_{\mathrm{SAE}}\) is strictly closer to the true \(V^\pi\) than \(\widehat V_{\mathrm{GRPO}}\) unless \(\mathbf V^\pi\) is well-approximated by \(\mathcal C\).
\item \textbf{\(V_E\) vs \(V_{\mathrm{SAE}}\).} There is no unconditional ordering: \(V_E\) trades variance for (potential) bias, while \(V_{\mathrm{SAE}}\) enforces structural consistency and reduces estimation-to-class error. Which estimator is closer to \(V^\pi\) depends on (i) how well \(\mathcal C\) matches \(\mathbf V^\pi\) and (ii) the bias magnitude of \(V_E\) within each subtree. 
\end{itemize}